\documentclass[10pt,twocolumn,letterpaper]{article}

\usepackage[pagenumbers]{iccv} 

\usepackage{enumitem}
\setitemize{noitemsep,topsep=0pt,parsep=0pt,partopsep=0pt}
\usepackage[normalem]{ulem}
\newcommand{\algoname}{GUIDE}
\usepackage{algorithm}
\usepackage{algorithmic}
\usepackage{pifont}
\definecolor{darkgreen}{rgb}{0.0, 0.5, 0.0}
\usepackage{array}
\usepackage{colortbl}
\usepackage{xcolor}
\usepackage{placeins}
\usepackage{bm}

\newcommand{\x}{{\bf x}}

\newcommand{\bPhi}{{\bm \Phi}}
\newcommand{\bpsi}{{\bm \psi}}
\newcommand{\bPsi}{{\bm \Psi}}

\newcommand{\bW}{{\bf W}}

\newcommand{\bomega}{{\bm \omega}}

\definecolor{iccvblue}{rgb}{0.21,0.49,0.74}
\usepackage[pagebackref,breaklinks,colorlinks,allcolors=iccvblue]{hyperref}

\def\confName{ICCV}
\def\confYear{2025}

\title{What's in a Latent? Leveraging Diffusion Latent Space for Domain Generalization}

\author{
Xavier Thomas$^{1}$\quad\quad Deepti Ghadiyaram$^{12}$\thanks{Corresponding author.} \\
$^1$Boston University \quad
$^2$Runway \\
{\tt\small {\{xthomas, dghadiya\}@bu.edu} }}

\begin{document}
\maketitle

\begin{abstract}
Domain Generalization aims to develop models that can generalize to novel and unseen data distributions. In this work, we study how model architectures and pre-training objectives impact feature richness and propose a method to effectively leverage them for domain generalization. Specifically, given a pre-trained feature space, we first discover latent domain structures, referred to as \emph{pseudo-domains}, that capture domain-specific variations in an unsupervised manner. Next, we augment existing classifiers with these complementary \emph{pseudo-domain} representations making them more amenable to diverse unseen test domains. We analyze how different pre-training feature spaces differ in the domain-specific variances they capture. Our empirical studies reveal that features from diffusion models excel at separating domains in the absence of explicit domain labels and capture nuanced domain-specific information. On $5$ datasets, we show that our very simple framework improves generalization to unseen domains by a maximum test accuracy improvement of over $\textbf{\text{4\%}}$ compared to the standard baseline Empirical Risk Minimization (ERM). Crucially, our method outperforms most algorithms that access domain labels during training. 
Code: \href{https://xthomasbu.github.io/GUIDE/}{https://xthomasbu.github.io/GUIDE}.
\end{abstract}



\section{Introduction}
\begin{figure}[!t]
    \centering
    {\includegraphics[width=1\linewidth]{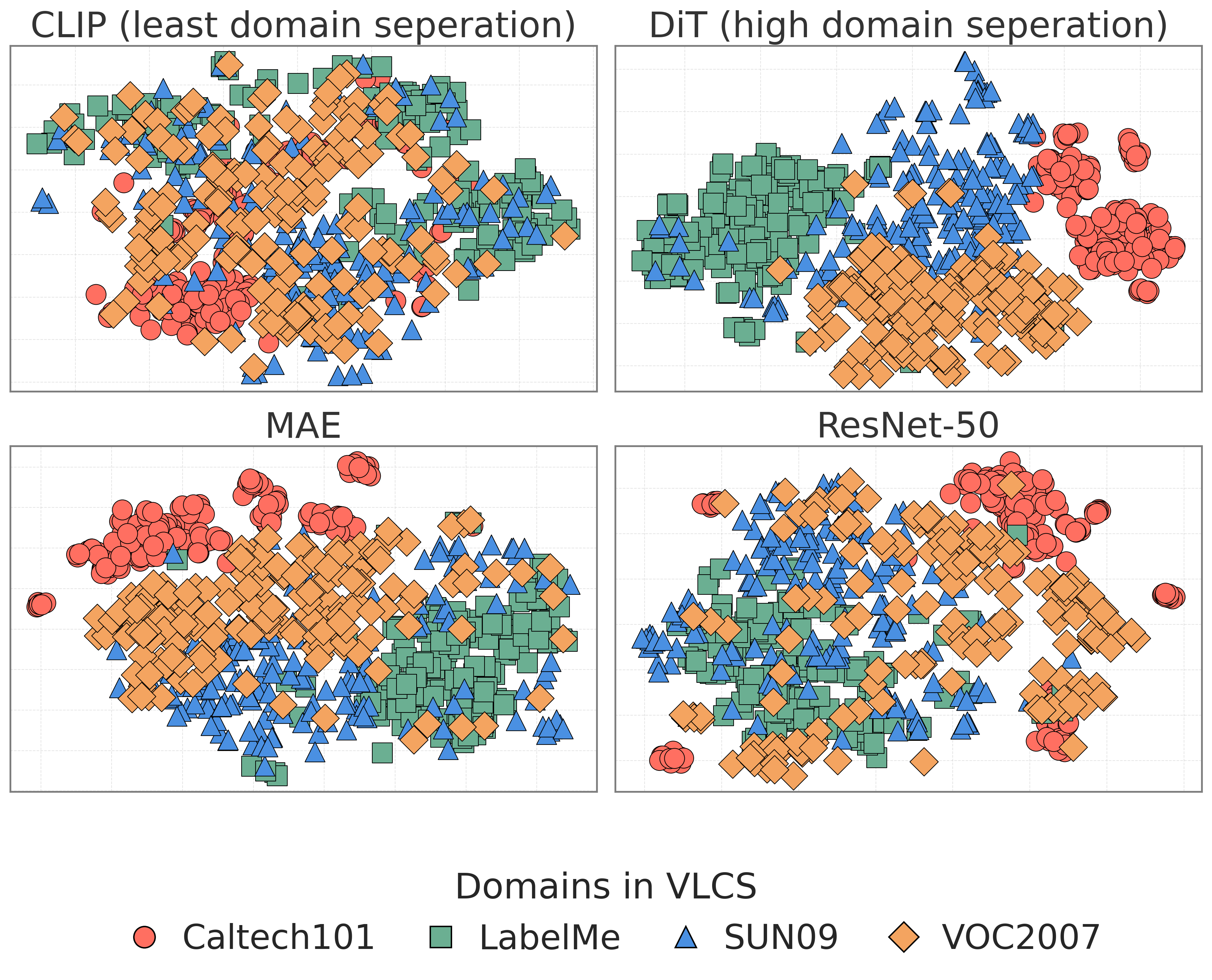}}
    \caption{\footnotesize{\textbf{T-SNE visualization of the latent space from different pre-training objectives:} CLIP~\cite{pmlr-v139-radford21a}, DiT~\cite{dit}, MAE~\cite{MAE}, ResNet-50~\cite{rn50} on the domain generalization benchmark VLCS~\cite{VLCS}. VLCS is curated from $4$ different datasets, thus dataset-specific biases like spatial composition and object size variations serve as different domains. Note how the diffusion features separate domains effectively, suggesting that latent domain structures can be captured without explicit supervision.} Best viewed in color.}\label{fig:main}
\end{figure} 
It is now a common practice to use models pre-trained on billion-scale data~\cite{rn50, MAE, stablediffusion, dit, Oquab2023DINOv2LR, pmlr-v139-radford21a, mahajan2018exploring} as defacto backbones for diverse downstream tasks~\cite{llava, eyeshut}. In order to make these large-scale models ``foundational,'' and offer rich feature representations, a variety of powerful pre-training strategies have been designed. Some of these objectives aim to eliminate the need for clean labeled data~\cite{simclr, deepcluster, clusterfit, swav, moco}, some reap the benefits from rich text representations by aligning them with corresponding visual signals~\cite{pmlr-v139-radford21a, ALIGN}, while others force models to build a more meaningful understanding of scenes by learning to predict large hidden regions of images~\cite{MAE}. Despite such tremendous progress, what exactly is captured in the underlying latent landscape remains an open question. This question becomes more challenging in diffusion models mainly due to their iterative global denoising objective.

This work aims to understand the feature landscape learnt from different pre-training models and objectives in the context of domain generalization. 
Robust generalization to unseen domains has been a long-standing goal in machine learning research~\cite{Blanchard2011GeneralizingFS, muandet2017kernel}, particularly in scenarios where collecting domain-specific data is infeasible or expensive. In such cases, models must learn to generalize without relying on explicit domain labels even during training~\cite{source_free_da}.
It has been established that most sophisticated models struggle when the test data distribution differs from that of training data~\cite{teterwak2024large,recht2019imagenet, taori} even in subtle ways, e.g., same visual scene but captured using different cameras, same patient but different brand imaging devices, same object but captured in different color schemes and so on.

We posit that the first step to make fundamental progress towards designing foundational models is to examine and interpret how current state-of-the-art models structure visual information and uncover their strengths and limitations. For instance, how are object, scene, and domain-specific variations internally encoded in a latent space? 
Do domain-specific traits manifest in distinct regions of the latent space or are they engulfed along with low- to mid-level scene and object level information? 

We study these questions in great detail in this work. Specific to the task of domain generalization, we analyze how different pre-training objectives and architectures influence the granularity of visual information captured in their feature space. Our key insight is that certain internal states of diffusion models effectively capture abstract information such as photographic styles, camera angles, and so on. Building on this insight, we first develop an unsupervised method for discovering latent domain structures. Next, we alter a standard domain generalization classification~\cite{ERM} pipeline with one key difference: we augment the classifier's representations with the discovered latent domain representations. We show through extensive empirical analysis that this simple tweak to the standard pipeline assists in training a model that generalizes well to unseen domains~\cite{ben2010theory}. While most prior works focus on leveraging a single feature space to design a universal model~\cite{coral, ganin2016domain, miro, liu2024latentdr, teterwak2304erm++}, we take an alterative approach and compliment existing classifier's features with \textit{domain rich} features and show that this auxiliary guidance makes the overall feature space more robust to unseen domains. Our framework dubbed \textbf{GUIDE}: \textbf{G}eneralization \textbf{u}sing \textbf{I}nferred \textbf{D}omains from Latent \textbf{E}mbeddings, offers a simple and effective method to ``guide" a given feature space to adapt better to unseen domains. 
We summarize our key contributions below:
\begin{itemize}[noitemsep,topsep=0pt]
\item We propose a method of \textbf{unsupervised pseudo-domain discovery} from frozen pre-trained feature spaces and use them to improve any model's ability to generalize to diverse domains, making it particularly useful in scenarios where domain labels used during training are unavailable or noisy  (Sec.~\ref{sec:diff_domain_gen}).
\item We \textbf{analyze different pre-training objectives and architectures} and investigate how they influence the structure of the feature latent landscape of both diffusion and conventional vision models (Sec.~\ref{sec:pretrain_obj}).
\item We shine light on \textbf{the ability of diffusion models to capture domain-specific information, such as photographic and artistic styles, texture variations}, and demonstrate their effectiveness to domain generalization (Sec.~\ref{sec:domainbed_results}). We obtain an average test accuracy improvement of $\mathbf{+2.6\%}$ on $5$ datasets, notably beating ERM~\cite{ERM} by $\mathbf{+4.3\%}$ on the TerraIncognita dataset~\cite{TerraInc}.
\end{itemize}

\section{Related Work}


{\bf Diffusion features for representation learning:} Diffusion models~\cite{sohl2015deep, ddpm} have significantly advanced image and video generation, prompting extensive exploration of their intermediate representations and their utility for diverse downstream tasks such as detection~\cite{diffusion_detect}, segmentation~\cite{label_seg, diffusion_segmentation}, classification~\cite{diffusion_classifier_23}, semantic correspondence~\cite{diff_hyper}, depth estimation~\cite{wu2023datasetdm, zhao2023unleashing}, and visual reasoning~\cite{diva}, showcasing their utility in both discriminative and generative domains. Recent studies~\cite{revelio, diff_hyper, voynov2023p+} demonstrate that features extracted across layers and timesteps encode rich semantic information, ranging from coarse patterns to fine-grained details. In this work, we analyze how the latent space of diffusion models captures class and domain-specific information and leverage these representations for the task of domain generalization.

\noindent {\bf Domain generalization:} First formalized in~\cite{Blanchard2011GeneralizingFS}, domain generalization is the challenging task of designing models capable of generalizing to unseen test domains. Various methods have been proposed to address this by learning domain-agnostic representations~\cite{nam2021reducing, huang2020self}, data or latent augmentation methods~\cite{hong2021stylemix, somavarapu2020frustratingly, li2021simple, liu2024latentdr}, and meta-learning~\cite{balaji2018metareg, bui2021exploiting}. 
Despite numerous advancements, most methods still under perform Empirical Risk Minimization (ERM) when evaluated rigorously~\cite{gulrajani2021in}, making it a very strong baseline. \citet{teterwak2304erm++} builds a stronger baseline by incorporating improved training strategies. 
\citet{matsuura2020domain}  learn a domain-invariant feature extractor by clustering samples into latent domains using style statistics from early convolutional layers, then applying adversarial learning to reduce domain distinctions.
\citet{bui2021exploiting} uses meta-learning and explicit domain labels to disentangle domain-invariant and domain-specific features, ensuring that the latter remains useful when adapting to new domains. The classifier then integrates both feature types for improved generalization.
~\citet{Dubey2021AdaptiveMF, thomas2021adaptivemethodsaggregateddomain} explore techniques to incorporate pseudo-domain information into classifiers to make them generalizable to unseen domains. Our work differs from these prior arts in several crucial ways: we leverage pre-trained models instead of learning a separate domain prototype network as in~\cite{Dubey2021AdaptiveMF}, utilize a more domain-rich feature space compared to~\cite{thomas2021adaptivemethodsaggregateddomain}, and do not rely on domain labels as in~\cite{bui2021exploiting, Dubey2021AdaptiveMF}.

\noindent \textbf{Diffusion models for domain generalization.} Prior works~\cite{yu2023distribution, hemati2023cross, huang2025domainfusion, hematibeyond} use text-to-image diffusion models as a data augmentation tool by generating diverse synthetic samples with variations that help models generalize better to unseen domains. However, these techniques rely on fine-tuning the diffusion model, expensive data augmentation steps, or access to the test data. By contrast, to the best of our knowledge, we are the first to investigate using frozen pre-trained diffusion features in an unsupervised manner for domain generalization.

\section{Approach}
First, we introduce the preliminaries of diffusion models (Sec. \ref{sec:diff_latent}) and domain generalization (Sec. \ref{sec:domain_gen}). Then, we present our two-step framework where we first learn pseudo-domain representations in an unsupervised manner and use them to adapt a classifier to unseen domains (Sec. \ref{sec:diff_domain_gen}). We stress that we \underline{do not have} domain label information during both training and test phases.
\subsection{Preliminaries on Diffusion Models}
\label{sec:diff_latent}
\label{sec:prelims}
Diffusion models~\cite{sohl2015deep, ddpm} are probabilistic generative models designed to learn the data distribution through an iterative denoising process. In the forward diffusion process, an image \(x\) is incrementally corrupted with noise ($\epsilon$) over \(T\) timesteps, resulting in a sequence of increasingly noisy images \(\{x_t\}_{t=1}^T\). In the reverse process of iterative denoising, a model \(\theta\), predicts the added noise \(\epsilon_\theta(x_t, t)\) at each timestep \(t\). 
Latent Diffusion Models~\cite{stablediffusion} (LDM) extend this framework by operating on a latent representation \(z\) of the image \(x\) instead of directly in its high-dimensional pixel space. This latent representation is obtained by mapping the image into a lower-dimensional space using a variational autoencoder~\cite{kingma2013auto} with an encoder \(E\) and a decoder \(D\). The diffusion process models the distribution of these lower-dimensional latent embeddings, enabling more efficient computation. The training objective is:
\[
L_{\text{LDM}} = \mathbb{E}_{E(x), t, \epsilon \sim \mathcal{N}(0, 1)} \|\epsilon - \epsilon_\theta(z_t, t)\|^2_2
\]

\subsection{Domain Generalization}
\label{sec:domain_gen}

Let $X$ and $Y$ be random variables denoting input and target labels respectively, and $\bPhi$ a feature extractor. In supervised learning, a predictor $f$ is learnt to map feature representations of inputs \(x \in X\), i.e., $\bPhi(x)$ to labels $y \in Y$, such that $f$ generalizes to unseen test samples. We denote this as $f(\bPhi(x)) \rightarrow y$. Domain generalization is an extension of supervised learning, where training data from multiple domains is available and the goal is to learn a predictor that performs well on samples from an \textit{unseen} test domain~\cite{Blanchard2011GeneralizingFS}. 

As in a conventional domain generalization framework, each domain $d$ is characterized by a probability distribution $P_d$ defined over $X$ and $Y$. The training dataset is constructed by sampling \(d^{tr}\) domains, denoted as \(\{P_d^{tr}\}_{d=1}^{d^{tr}}\), and collecting \(n_d\) labeled points from each domain, forming the dataset $\bigcup_{d=1}^{d^{tr}} \{(x_i^d, y_i^d)\}_{i=1}^{n_d}$. The unseen test domain distribution is denoted as \(P_d^{te}\), from which \(n_T\) unlabeled points \(\{x_i^{d^{te}}\}_{i=1}^{n_T}\) are sampled during evaluation. 

One popular approach for domain generalization is to learn a universal classifier on all training samples~\cite{ERM} that is agnostic to the underlying domains. However, this algorithm makes a strong assumption that all training samples are drawn from a single, unified distribution and minimizes the average risk across them. Though simple and effective, this may not guarantee good performance, especially when the test domain lies further from the assumed unified distribution or when the training domains themselves have a very high variance~\cite{Dubey2021AdaptiveMF}. To address this, motivated by findings in~\cite{Dubey2021AdaptiveMF, thomas2021adaptivemethodsaggregateddomain, matsuura2020domain, bui2021exploiting} which leverage domain-specific representations, we complement input features with these representations. We hypothesize that augmenting input features with rich, \textit{complementary} information about (pseudo) domains would make the overall feature space more robust to diverse domain variations.

\noindent \textbf{Control experiment using ground truth domain labels:} To validate the above hypothesis, we conduct the following control experiment. We assume access to ground truth domain labels, cluster diffusion features explicitly into each domain, and compute cluster centroids. Next, we augment the input features ($\bPhi(x)$) by concatenating them with the cluster centroids and train a classifier on them. On a popular domain generalization benchmark OfficeHome~\cite{OfficeHome}, we achieve a boost of $\textbf{3\%}$ over the strongest baseline. We acknowledge that the number of pseudo-domains we learn per dataset in {\algoname} (Sec.~\ref{sec:diff_domain_gen}) is different from the true domains present in each dataset. Yet, this controlled setup highlights that augmenting a feature space with domain-specific representations from \textit{seen} domains yields an overall generalizable feature space for \underline{unseen} domains.

Though the standard domain generalization framework assumes access to domain labels during training, in certain applications, this information may be unavailable or incorrect. Thus, we design a robust algorithm to learn this complementary ``pseudo'' domain information, described next.

\subsection{Adaptive Domain Generalization}
\label{sec:diff_domain_gen}
\begin{figure}[!t]
    \centering
    {\includegraphics[width=1\linewidth]{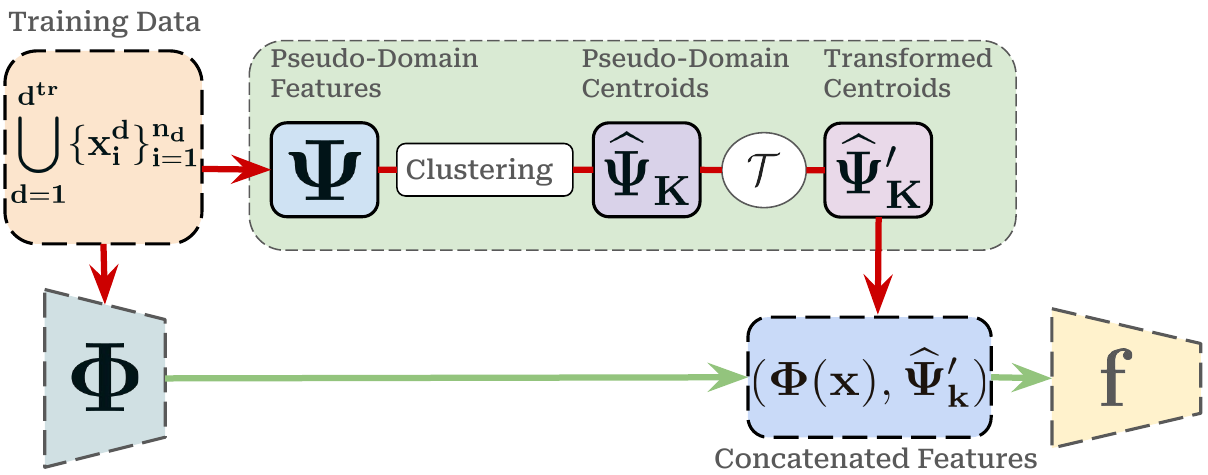}
    \caption{\footnotesize{\textbf{Training Pipeline.} The \textcolor[rgb]{0,0.5,0}{green-shaded region} represents the clustering and transformation step. \textcolor[rgb]{0,0.6,0}{Green solid arrows} indicate gradient flow, while \textcolor[rgb]{0.9,0,0}{red arrows} represent non-gradient operations. The feature extractor $\bPsi$ first clusters samples to compute the pseudo-domain centroids. The transformation function $\mathcal{T}$ then transforms these centroids to the latent space of $\bPhi$, producing transformed pseudo-domain centroids, which are concatenated with the features from $\bPhi$, and sent to the classifier.}}
    \label{fig:pipeline}
    \vspace{-0.8em}}
\end{figure} 
\begin{table*}[t]
\centering
\resizebox{0.95\textwidth}{!}{ 
\begin{tabular}{lcc}
\hline
\textbf{Dataset}      & \textbf{Domain Shift Type}              & \textbf{Example}       \\ \hline
PACS~\cite{PACS}       & Style and texture variations                & Monochromatic ``Sketch", color-rich ``Art Painting"      \\
VLCS~\cite{VLCS}       & Dataset-specific biases           & Spatial composition and object size variations (e.g., objects in ``Caltech" are centered)       \\
TerraIncognita~\cite{TerraInc} (TI) & Environmental and background shifts       & Location-specific foliage and terrain variations \\
OfficeHome~\cite{OfficeHome} (OH) & Low-level style differences     & ``clipart" has bold edges, ``real" with softer, natural edges \\
DomainNet~\cite{domainnet} (DN)     & Style, texture, and complexity differences & ``Quickdraw" has simplistic and often incomplete outlines, ``Sketch" features more refined strokes with shading \\ 
Synth-Artists         & Synthetic artistic styles         & Variations in artist techniques, color schemes, and brushwork \\ 
Synth-Photography     & Synthetic photographic styles    & Changes in lighting, contrast, focus \\ \hline
\end{tabular}
}
\caption{\footnotesize\textbf{Overview of domain shifts in each dataset}, including low-level and global photographic style variations, environmental, and dataset-specific biases. Example images for each dataset in suppl. material.}
\label{tab:domain_shifts}
\end{table*}

\textbf{Learning pseudo-domain representations:} 
In the absence of true domain labels, we adopt an unsupervised method called Kernel Mean Embeddings (KME)~\cite{Blanchard2011GeneralizingFS, muandet2017kernel} to capture key statistical properties of a domain. KMEs offer an efficient way to summarize and represent a probability distribution into a single, representative feature vector. In our case, given the probability distributions of the training domains \(\{P_d^{tr}\}_{d=1}^{d^{tr}}\), we use the feature extractor \(\bPsi\) to compute feature representations for samples drawn from each \(P_d^{tr}\). Then, we apply K-Means++ clustering and obtain $K$ clusters as a way to capture the underlying domain structures. Given we lack information about the true number or nature of the underlying domains during training in our setup, we refer to these clusters as \textit{pseudo} domains. The centroid of each cluster $\widehat\bPsi_{k}$, for $k \in K$ is used as the compact representation of each pseudo-domain. Finally, we assign each training sample $x$ to its nearest cluster, such that it's pseudo-domain feature representation is $\widehat\bPsi_{x}$ = $\widehat\bPsi_{k}$, the centroid of the corresponding pseudo domain. We study the impact of different feature extractors $\Psi$ in Sec.~\ref{sec:impl} and \ref{sec:pretrain_obj}. We show how clustering smooths out any noise or sample-specific variations and creates more stable (pseudo) domain representations in Sec.~\ref{sec:domainbed_results}.

\noindent \textbf{Leveraging pseudo-domain representations:} 
We take inspiration from ERM~\cite{ERM} and learn a single universal classifier on all training domains, with one key difference: we augment each input feature vector with it's corresponding pseudo-domain representation. Specifically, we first apply a transformation function on the pseudo-domain representations to bring the latent manifold of $\bPsi$ closer to $\bPhi$ to mitigate feature domain drift, 
i.e., $\mathcal{T}:\bPsi \mapsto \bPhi$. Then, we concatenate the input feature vector $\bPhi(x)$ with it's corresponding pseudo-domain representation $\mathcal{T}(\widehat\bPsi_k)$ during training, to learn a ``domain-adaptive" classifier (as introduced in ~\citet{Dubey2021AdaptiveMF}). At test time, we first process the input through $\bPsi$, then assign it to the nearest cluster centroid learned during training, and finally apply $\mathcal{T}$ before passing through the classifier. \textit{We stress that in our setup, we do not assume access to domain information during training and make no assumptions about the test domains.} 

\section{Experiments}

We outline the implementation details and training setup for {\algoname} in Sec~\ref{sec:impl}, followed by a detailed analysis of the capability of different feature extractors ($\bPsi$) in capturing domain-specific information to augment class-specific features ($\bPhi$) in Sec~\ref{sec:pretrain_obj}. We empirically show how our approach leads to a more domain generalizable classifier on unseen test domains and the role of clustering in Sec.~\ref{sec:domainbed_results}.


\subsection{Implementation Details} 
\label{sec:impl}

\begin{table}[!t]
\centering
\resizebox{0.45\textwidth}{!}{%
\begin{tabular}{lcc}
\toprule
\textbf{Model} & \textbf{Source} & \textbf{Feature Dimension} \\
\midrule
ResNet-50      & Global Average Pooling (GAP) at layer 49 & 2048  \\
CLIP (ViT-L/14) & CLS token & 1024  \\
DINOv2 (ViT-L/14)  & Mean over patch tokens & 1024  \\
MAE (ViT-L/14) & Mean over patch tokens & 1024  \\
SD-2.1         & Mean over channels of $up\_ft1$ layer~\cite{revelio} & 1280  \\
DiT-XL/2-512   & Mean over tokens of block 14~\cite{revelio} & 1152  \\
\bottomrule
\end{tabular}}
\caption{\footnotesize \textbf{Feature extraction details} from each model. SD-2.1 features are conditioned on an empty text prompt.}
\label{tab:feature_extraction}
\end{table}

\textbf{Datasets:} We conduct our experiments on $7$ datasets, summarized in Table~\ref{tab:domain_shifts}. Five of these datasets (PACS, VLCS, TerraIncognita, OfficeHome, DomainNet) are part of the DomainBed~\cite{gulrajani2021in} test bed. We present details of Synth-Artists, and Synth-Photography in Sec.~\ref{sec:appl}.

\noindent \textbf{Training Setup:} We use the default hyper parameter settings from DomainBed~\cite{gulrajani2021in}: a batch size of $32$ per domain, learning rate of $5 \times 10^{-5}$, number of steps as $5001$, no dropout in the backbone model, and a weight decay of $0$ on $1$ A6000 GPU. We report test accuracies using the leave-one-domain-out cross-validation methodology~\cite{gulrajani2021in}, and average the results obtained over 3 trial seeds. \\
\textbf{Choice of $\bPhi$:} We use ResNet-50~\cite{rn50}, initialized with AugMix~\cite{hendrycks2019augmix} pre-trained weights as in DomainBed~\cite{gulrajani2021in}.

\noindent \textbf{Choice of $\bPsi$:} We study the feature spaces from several vision encoders with varied pre-training objectives: cross-entropy loss-based  ResNet~\cite{rn50}, contrastive loss-based CLIP~\cite{pmlr-v139-radford21a}, a distillation-based loss in DINOv2~\cite{Oquab2023DINOv2LR}, and reconstruction of masked patches loss-based MAE~\cite{MAE}. We further study two diffusion model architectures: the convolutional UNet-based~\cite{unet} Stable Diffusion 2.1 (SD-2.1)~\cite{stablediffusion} and transformer-based DiT-XL-2-512 (DiT)~\cite{dit}. Though the underlying pre-training objective is the same for diffusion models, we aim to study the influence of the underlying diffusion architecture on the learnt feature landscape. We provide details on the layers from which the features are extracted in Table.~\ref{tab:feature_extraction}.\\
\textbf{Choice of $\mathcal{T}$ and cluster refinement schedule:} To adapt the pseudo-domain representations as $\bPhi$ evolves during training, we define $\mathcal{T}$ (Sec~\ref{sec:diff_domain_gen}) as a radial basis function (RBF) kernel ridge regressor (more in suppl. material). RBF kernels are well-known for their ability to model non-linear, distance-based relationships and have been effectively used to align second-order statistics between source and target distributions~\cite{Zhang2018MCA}. In our approach, $\mathcal{T}$ maps the centroid $\widehat{\bPsi}_{k}$ of a given pseudo-domain $k$ to the mean of the features $\bPhi(x)$ of the samples  belonging to cluster $k$. We employ a logarithmic schedule~\cite{thomas2021adaptivemethodsaggregateddomain} to periodically apply $\mathcal{T}$ on $\widehat{\bPsi}_{k}$, 
starting with frequent updates and progressively reducing their frequency and thus the overall computational overhead. We note that clustering is done only once on the static $\bPsi$-feature space, but the refinement follows a logarithmic schedule. 

\noindent \textbf{Number of Pseudo-Domains:} For {\algoname}, the number of clusters ($K$) is the sole hyper-parameter. We follow a simple heuristic from~\citet{thomas2021adaptivemethodsaggregateddomain} to determine this: $K = \text{max} \bigl( \{1, 3, 5\} \times n_c, 200\bigr)$, where $n_c$ represents the number of classes in the dataset. The upper-bound of $200$ clusters helps prevent over-clustering. The number of clusters that yields the best test accuracy for each domain is used to report the scores in Table~\ref{tab:domainbed}.

\noindent\textbf{Evaluation of domain separability:} With a motive to measure \textit{expressivity}~\cite{Dubey2021AdaptiveMF} of the underlying pseudo-domain representations, we measure normalized mutual information (NMI) as done in prior works~\cite{matsuura2020domain, thomas2021adaptivemethodsaggregateddomain}. In our setup, let \( U \) and \( V \) be random variables that denote pseudo domain labels and ground truth domain (or class) labels. NMI is defined as:
\[
\text{NMI}(U, V) = \frac{2 \cdot I(U, V)}{H(U) + H(V)},
\]
where \( I(U, V) \) is the mutual information between \( U \) and \( V \) and \( H(U) \), \( H(V) \) their respective entropies. NMI measures how well the discovered clusters match the ground truth domain or class labels. In our setup, a feature space that yields clusters having high domain-NMI score is an ideal candidate to complement existing class-specific features. 
\subsection{Underlying Domains in Each Dataset}
\label{sec:domain_shift_sum}
We begin by summarizing the types of domain shifts present in the datasets we study (described in Table~\ref{tab:domain_shifts}). PACS~\cite{PACS} image dataset captures $7$ object categories and $4$ domains: real-world photos, art paintings, cartoons, and sketches. Thus, the domains have stark visual distinctions driven by both global and local changes such as shapes, colors, and edges. VLCS~\cite{VLCS} is curated from different datasets, making dataset-specific biases such as spatial composition and object size variations as different domains. OfficeHome~\cite{OfficeHome} similar to PACS also has images belonging to four domains: artistic, clip-art, product catalog, and real-world images. Thus, while there is some overlap in the underlying structural characteristics of the objects across domains, the domain shifts primarily involve style differences such as variations in texture, color, and outlines. TerraIncognita~\cite{TerraInc} consists of images taken from different camera trap locations, and each camera serves as a domain. Thus, the domain shifts are driven by physical environmental aspects such as variations in foliage density, terrain patterns, and spatial patterns of vegetation. 
DomainNet~\cite{domainnet} is composed of six domains such as quick-draw, infographic, real images, and so on, and exhibits a broader range of domain shifts than PACS, spanning both coarse and fine-grained variations. For example, the ``quickdraw" domain consists of simple, rough sketches, while ``sketch" has more detailed drawings with shading and varied strokes, showing style differences. By contrast, ``real" domain captures fully detailed images, indicating shifts of varied granularities between different domains.
\begin{table}[!t]
\centering
\resizebox{0.45\textwidth}{!}{%
\begin{tabular}{lcccccc}
\toprule
\textbf{Dataset} & \textbf{DiT} & \textbf{SD-2.1} & \textbf{MAE} & \textbf{CLIP} & \textbf{DINOv2} & \textbf{RN50} \\
\midrule
PACS             & \textbf{0.85}            & 0.82           & 0.71            & 0.54             & 0.55             & 0.59             \\
VLCS             & \textbf{0.58}            & 0.26           & 0.20            & 0.01             & 0.05             & 0.22             \\
TerraInc   & 0.22            & \textbf{0.55}           & 0.21           & 0.01            & 0.01             & 0.25            \\
OfficeHome   & 0.25            & 0.28           & 0.10           & 0.12           & \textbf{0.38}             & 0.08           \\
DomainNet   & \textbf{0.54}            & 0.51           & 0.52           & 0.32           & 0.47             & 0.46           \\
Synth-Artists & \textbf{0.89}           &   0.86         & 0.75            & 0.25             & 0.34             & 0.63           \\
Synth-Photography & 0.35           & \textbf{0.43}           & 0.31            & 0.17             & 0.23             & 0.33           \\
\bottomrule
\end{tabular}}
\caption{\footnotesize \textbf{Comparison of domain NMI scores across datasets}. The highest domain NMI score depends both on the type of pre-training feature space and the underlying domain shifts in the dataset as noted in Sec~\ref{sec:pretrain_obj}. We note that inherent domain label noise can impact domain NMI scores. Thus, NMI is more valuable when used as a relative measure rather than an absolute indicator of domain separability.}
\label{tab:dataset_comparison}
\end{table}
\begin{table}[!t]
\centering
\resizebox{0.40\textwidth}{!}{%
\begin{tabular}{lcccccc}
\toprule
\textbf{Dataset} & \textbf{DiT} & \textbf{SD-2.1} & \textbf{MAE} & \textbf{CLIP} & \textbf{DINOv2} & \textbf{RN50} \\
\midrule
PACS             & 0.08            & 0.08          & 0.11            & 0.05             & 0.15             & 0.29             \\
VLCS             & 0.12            & 0.15          & 0.17            & 0.01             & 0.11             & 0.39             \\
TerraInc   & 0.32            & 0.35           & 0.32           & 0.01            & 0.16             & 0.30            \\
OfficeHome   & 0.16            & 0.22          & 0.28           & 0.10            & 0.23             & 0.59            \\
DomainNet   & 0.16             &0.20           &0.22            &0.13             &0.19              &0.36             \\
\bottomrule
\end{tabular}}
\caption{\footnotesize \textbf{Comparison of class NMI scores across datasets.} In order to choose auxiliary features for domain separation, a feature space that yields lower class NMI score along with high domain NMI is desirable, i.e. the latent space should favor grouping domains over object classes. Note that Synth-Artists and Synth-Photography datasets are omitted here as they do not have predefined class labels. }
\label{tab:class_nmi_dataset_comparison}
\end{table}
\subsection{Effect of the Choice of \texorpdfstring{$\mathbf{\Psi}$}{\textbf{Psi}} on Domain Separation}
\label{sec:pretrain_obj}
Next, we study how different pre-training objectives affect the separation of domain-specific signals using \textbf{domain NMI} (\(\uparrow\)) (introduced in Sec.~\ref{sec:impl}), which measures how well domains are separated in the latent space. We acknowledge that all models are of varied architectural complexities, trained on very different datasets, thereby making it nonviable to concretely isolate the cause of performance discrepancies in domain separation. Nevertheless, we believe our below analysis is valuable to understand the semantic information captured by different pre-training objectives. \\
\noindent \textbf{ResNet-50~\cite{rn50} (RN50)} is pre-trained on ImageNet~\cite{deng2009imagenet} using a discriminatory cross-entropy classification loss. Consequently, the feature space evolves to aid object discrimination, making samples from the same class cluster together across domains. This is evident in the relatively low domain NMI scores (Table~\ref{tab:dataset_comparison}) and high class NMI across all datasets (Table~\ref{tab:dataset_comparison}), e.g., a class NMI of $0.29$ compared to $0.08$ by DiT features on the PACS dataset. 

\noindent \textbf{CLIP~\cite{pmlr-v139-radford21a}} is pre-trained on internet-scale, noisy image-text pairs using a contrastive loss that aligns images with their textual descriptions in a joint embedding space. This objective prioritizes high-level semantic similarity, making CLIP's feature space more representative of global semantics and overall context of the image instead of object-specific details. Consequently, images of the same object may not form tight clusters if their captions emphasize different contextual attributes (e.g. ``a dog on a beach" vs. ``a golden retriever indoors"). Thus, CLIP, though rich in broader contextual semantics, yields low class and domain NMI scores across all datasets in (Tables~\ref{tab:dataset_comparison} and~\ref{tab:class_nmi_dataset_comparison}).


\noindent \textbf{DINOv2~\cite{Oquab2023DINOv2LR}} is a self-supervised vision transformer trained by aligning representations between a student and a teacher network, across global and local crops of an image. This encourages the model to capture primarily low-level features, while also capturing global relationships to some extent~\cite{Oquab2023DINOv2LR, jiang2023clip, eyeshut}. By enforcing consistency across augmentations, DINOv2 preserves low-level features that remain invariant to these transformations. Thus DINO-v2 features are particularly effective for datasets like OfficeHome (domain NMI of $0.38$ in Table~\ref{tab:dataset_comparison}), where domain shifts are driven by low-level style differences such as bold outlines in the ``clipart" domain vs softer, natural edges in the ``real" domain (example images in suppl. material). By contrast, DINOv2 performs poorly on VLCS (domain NMI of $0.05$ in Table~\ref{tab:dataset_comparison}), likely due to its over-reliance on low-level features, making it less effective at capturing high-level dataset-specific biases in VLCS, such as differences in spatial composition and object size variations.

\noindent \textbf{Masked Autoencoders~\cite{MAE} (MAEs)} are pre-trained using a masking objective, where the model learns to reconstruct locally masked patches of an image. We conjecture that by reconstructing small, local patch details, MAE's pre-training objective may introduce a strong locality bias, and fail to capture global image context, as studied in~\cite{zhenda2022revealing, liang2022supmae}. We hypothesize that this lack of global understanding limits the capability of MAEs to offer complimentary domain-specific representations. This is evident in their relatively high class-NMI scores (as seen in Table~\ref{tab:class_nmi_dataset_comparison}) and low domain-NMI scores (as seen in Table~\ref{tab:dataset_comparison}) across most datasets. 
However, MAEs achieve relatively high domain NMI scores on PACS ($0.71$) and DomainNet ($0.52$) leveraging the visual information from local details such as textures, shading, and brushstrokes.
We note a similar trend with DINOv2 which also captures rich local features. This may explain why both models perform better in separating domains driven by low-level visual variations (Table~\ref{tab:dataset_comparison}). However, MAEs perform poorly on TerraIncognita despite its reliance on local features. Unlike PACS, we think that the domain shifts in TerraIncognita require an understanding of both local and global spatial understanding (e.g., vegetation density, terrain patterns), potentially leading to lower domain NMI.

\noindent \textbf{Conclusion:} This indepth analysis indicates that comprehending different pre-training objectives is essential to maximize profit from their latents for domain separation.

\begin{figure}[!t]
    \centering
    {\includegraphics[width=0.97\linewidth]{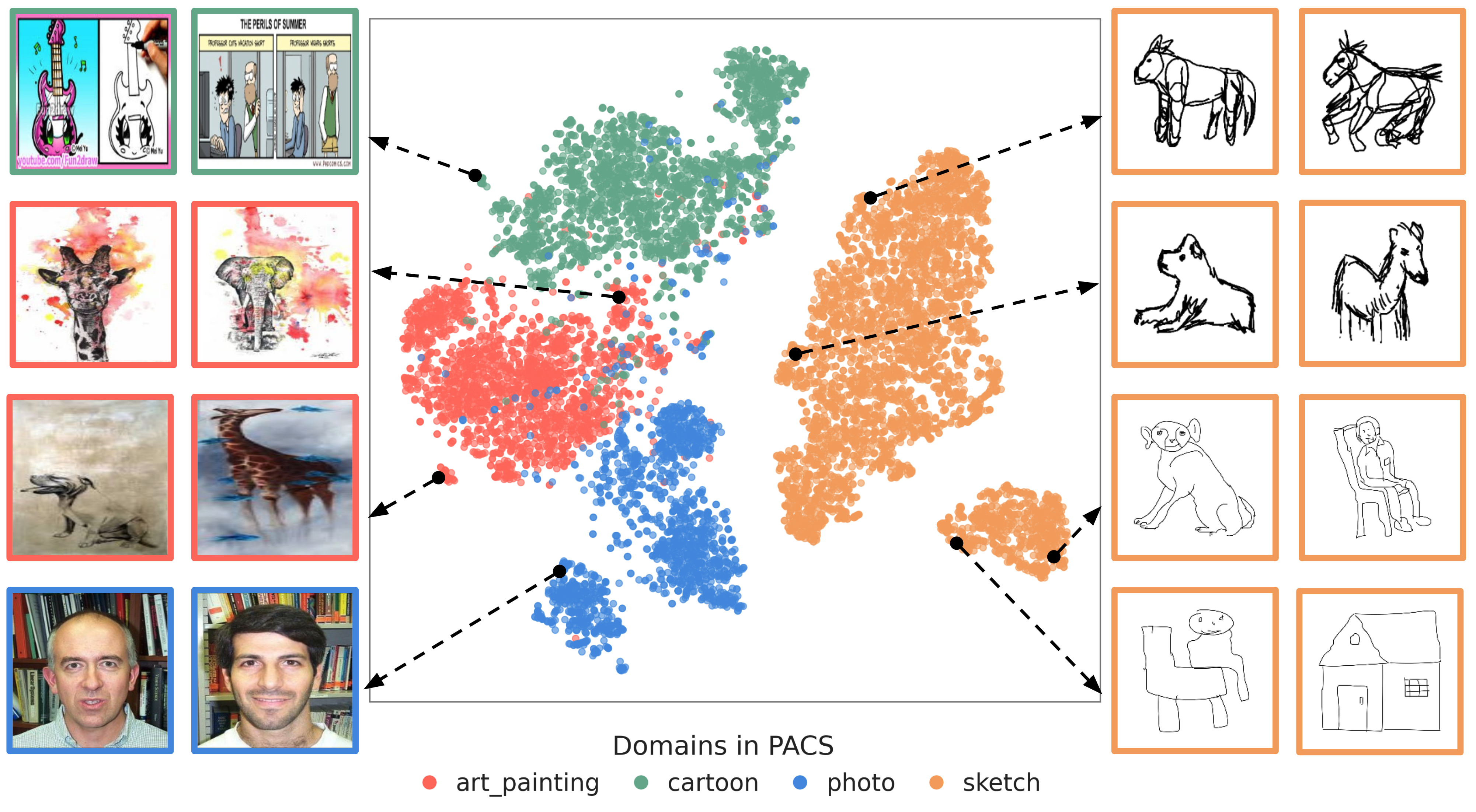}}
    \caption{\footnotesize{\textbf{T-SNE visualization of how pseudo-domains are clustered together in the latent space of DiT for PACS.} Note how the sketch domain forms distinct clusters, with light and dark pencil strokes mapped to separate regions in the latent space. Best viewed in color.}}
    \label{fig:pseudo_image}
\end{figure} 
\subsection*{Diffusion models for domain separation} 
\begin{figure}[!t]
    \centering

    \begin{minipage}{0.95\linewidth}
        \begin{minipage}{0.18\linewidth} 
            \raggedright \scriptsize (a) Animal portraits
        \end{minipage}%
        \hspace{0.1em} 
        \begin{minipage}{0.80\linewidth}
            \centering
            \includegraphics[width=\linewidth]{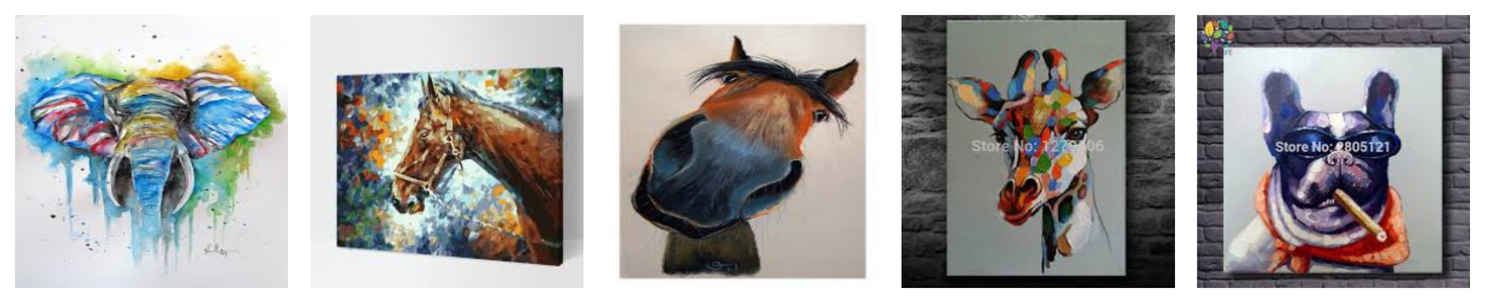}
        \end{minipage}
    \end{minipage}

    \vspace{-0.5em}

    \begin{minipage}{0.95\linewidth}
        \begin{minipage}{0.18\linewidth} 
            \raggedright \scriptsize (b) Oil paintings
        \end{minipage}%
        \hspace{0.1em}
        \begin{minipage}{0.80\linewidth}
            \centering
            \includegraphics[width=\linewidth]{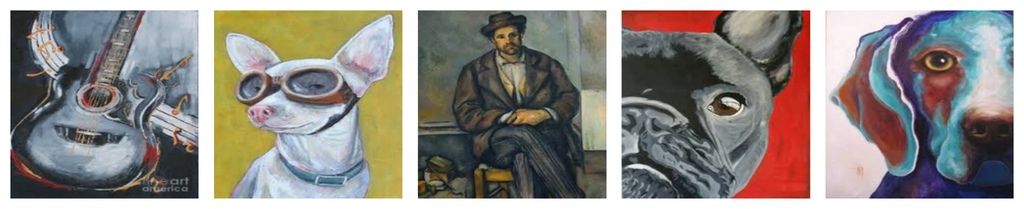}
        \end{minipage}
    \end{minipage}

    \vspace{-0.5em}

    \begin{minipage}{0.95\linewidth}
        \begin{minipage}{0.18\linewidth} 
            \raggedright \scriptsize (c) Similar color schemes
        \end{minipage}%
        \hspace{0.1em}
        \begin{minipage}{0.80\linewidth}
            \centering
            \includegraphics[width=\linewidth]{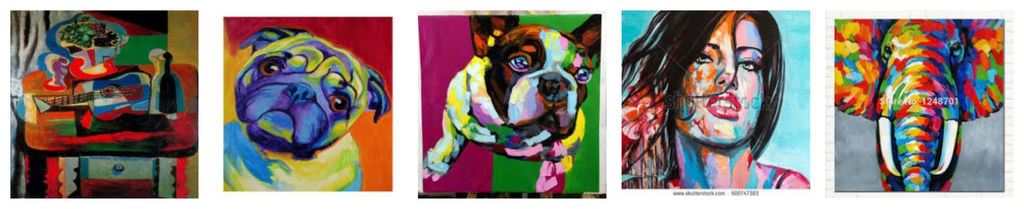}
        \end{minipage}
    \end{minipage}

    \caption{\footnotesize \textbf{Pseudo-domains captured in the diffusion latent space of DiT on PACS.} The clusters group images based on nuanced style-specific variances rather than class-specific variances.}
    
    \label{fig:pseudo_domain_cluster}
\end{figure}

Next, we focus exclusively on diffusion architectures and closely study the impact of some of their architectural design choices on domain separation. As discussed in Sec.~\ref{sec:prelims}, during diffusion model pre-training, noise added to an image is iteratively removed using pixel reconstruction loss. Recent studies~\cite{qian2024boosting, park2023understanding} have indicated that this makes the model first capture broad structural patterns before encoding finer details. We hypothesize that this \textit{implicit} hierarchical feature learning indirectly induced by the denoising objective enables diffusion models to encode global structures and fine-grained variations, assisting faithful image reconstruction. Moreover, since the generative objective is entirely agnostic to class labels, we posit that there is no incentive to group features based on class-discriminative signals. Perhaps this lack of class-driven objective allows domain-specific variations to emerge more prominently in the latent space. This is reflected in Table~\ref{tab:dataset_comparison} where we observe that diffusion features achieve high domain NMI scores across most datasets compared to their non-diffusion counterparts. Figures~\ref{fig:pseudo_domain_cluster}, and~\ref{fig:pseudo_image} further reinforces this observation and illustrates how different clusters (pseudo-domains) in the diffusion latent space capture domain-specific class-agnostic variations. 

Within the family of diffusion feature space, we now inspect if transformer based DiT and U-Net based SD-2.1 behave differently for the task of domain separation. We acknowledge that both models are trained on very different datasets which makes this analysis more challenging.

\noindent \textbf{DiT}~\cite{dit}: Following the analysis in~\citet{revelio}, we extract features from the 14th (out of 28) block of the transformer architecture of the DiT model, at timestep $t$=50 (more in suppl. material). As noted in~\cite{revelio}, by attending to the entire image, DiT’s self-attention mechanism effectively captures global context, making it capable at distinguishing high-level semantics and stylistic differences (e.g, pencil sketches vs paintings). This proves advantageous on datasets like PACS~\cite{PACS} which comprises domains with varied global context (detailed in Sec.~\ref{sec:domain_shift_sum}), where DiT achieves the highest domain NMI of $0.85$ (Table~\ref{tab:dataset_comparison}).

\noindent \textbf{SD-2.1}~\cite{stablediffusion}.  We extract features from the second upsampling layer of the U-Net (denoted as \texttt{up\_ft:1}) in SD-2.1 at timestep $t$=50 (more in suppl. material). As noted in~\cite{revelio}, these features are rich in fine-grained visual information, with convolutional-based U-Net~\cite{unet} of SD-2.1~\cite{stablediffusion} capturing local spatial information~\cite{plugandplay, revelio}. As a result, we observe that SD-2.1 and DiT exhibit complementary strengths. This is particularly evident on the TerraIncognita dataset, where SD-2.1 achieves the highest domain NMI of $0.55$ (Table~\ref{tab:dataset_comparison}). This is likely due to SD-2.1's ability to capture fine-grained spatial features such as foliage density and terrain patterns, which define the domain shifts for this dataset (as described in Sec~\ref{sec:domain_shift_sum}). By contrast, we find that DiT struggles with domain separation on TerraIncognita, achieving a lower domain NMI of $0.22$. On the other hand, on VLCS~\cite{VLCS} where each domain represents dataset-specific biases described in Sec.~\ref{sec:domain_shift_sum}, we note that DiT (domain NMI: $0.58$) outperforms SD-2.1 (domain NMI: $0.26$). This highlights DiT's strength to capture global context. Interestingly, SD-2.1’s \texttt{bottleneck} layer achieves a higher domain NMI of $0.45$ compared to \texttt{up\_ft:1}'s score of $0.26$. This aligns with the findings from~\citet{revelio} that U-Net's \texttt{bottleneck} layer captures coarser, more global features, compared to \texttt{up\_ft:1}.

From Table~\ref{tab:dataset_comparison}, we observe that OfficeHome~\cite{OfficeHome} proves to be challenging for both DiT (domain NMI: $0.25$) and SD-2.1 (domain NMI: $0.28$). Upon inspection, we found that samples from the ``real" domain visually look similar to those from both ``product" and ``art" in the feature spaces (suppl. material for visual examples), potentially contributing to low domain separation. On DomainNet~\cite{domainnet}, we observe moderate domain NMI scores for all pre-training objectives (except for CLIP, as discussed in Sec~\ref{sec:pretrain_obj}), with DiT achieving the highest score of $0.54$, in Table~\ref{tab:dataset_comparison}. We attribute this to the diverse nature of domain shifts in DomainNet, which include both high- and low-level variations (described in Sec.~\ref{sec:domain_shift_sum}). We believe that this variability makes it challenging for models to fully leverage their distinct strengths, as no single model seems to effectively capture all domain-specific characteristics.

\noindent \textbf{Conclusion:} This analysis reveals that for the same pre-training objective (diffusion denoising), the underlying architecture and the specific layer for feature extraction plays a crucial role in shaping the latent space, thereby performance on the downstream tasks. 

\subsection{Domain Generalization Performance}
\label{sec:domainbed_results}

In this section, we compare {\algoname} against prior domain generalization methods and examine the impact of different feature extractors ($\bPsi$) in capturing domain-specific information to enhance classification performance. 

\begin{table}[!t]
\centering
\resizebox{0.45\textwidth}{!}{%
\begin{tabular}{lcccccc|c}
\toprule
\textbf{Dataset} & \textbf{DiT}  & \textbf{SD-2.1} & \textbf{RN50} & \textbf{CLIP} & \textbf{DINOv2} & \textbf{MAE} & \textbf{ERM} \\
\midrule
VLCS     & \textbf{78.5}   & 77.0    & 76.3  & 76.8 & 77.3  & 76.4  & 76.6 \\
PACS   & \textbf{87.1}    & 86.9   & 84.8  & 84.7  & 84.9  & 84.6  & 83.8 \\
OH     & 68.4   & \textbf{68.6}    & 65.7  & 64.6 & 68.3  & 65.2  & 67.2 \\
TI     & 48.2   & \textbf{51.3}    & 49.8  & 47.4  & 48.4  & 50.2  & 47.0 \\
\midrule
\rowcolor{gray!10}Avg & 70.6  & \textbf{71.0}    & 69.1  & 68.4  & 69.7  & 69.1  & 68.7 \\
\bottomrule
\end{tabular}}
\caption{\footnotesize \textbf{Domain generalization performance on PACS and TerraIncognita (TI)}. The pseudo-domain representations obtained from the latent space of diffusion models provide the highest gains in accuracy, while those from CLIP yield minimal accuracy gains. 
}
\label{tab:domain_gen_spaces}
\end{table}

\noindent \textbf{Choice of $\Psi$ on domain generalization:} Building on our findings in Sec.~\ref{sec:pretrain_obj}, we test the utility of different feature spaces for domain separation and generalization against ERM~\cite{ERM}, a strong baseline that has been shown by~\citet{gulrajani2021in} to outperform many domain generalization algorithms. We evaluate on the DomainBed test suite, which comprises PACS~\cite{PACS}, VLCS~\cite{VLCS}, OfficeHome~\cite{OfficeHome}, TerraIncognita~\cite{TerraInc}, and DomainNet~\cite{domainnet}. From Table~\ref{tab:domain_gen_spaces}, we note that diffusion features consistently outperform their non-diffusion counterparts on all datasets. Notably, DiT and SD-2.1 achieve highest accuracy while the rest show only marginal gains over ERM. CLIP seems to yield minimal gains on average on this task limiting its ability to be used ``as is.''  GUIDE-DiT yields an average accuracy improvement of $\textbf{1.9\%}$ over ERM and performs best on VLCS ($\textbf{+1.9\%}$) and PACS ($\textbf{+3.3\%}$). On the other hand, GUIDE-SD-2.1 outperforms on TerraIncognita, beating ERM by $\textbf{+4.3\%}$. These results are inline with the analysis and domain NMI scores in Table~\ref{tab:dataset_comparison}.

\begin{table}[!t]
\centering
\resizebox{0.47\textwidth}{!}{%
\begin{tabular}{>{\centering\arraybackslash}m{1.55cm}|>{\centering\arraybackslash}m{1.15cm}| lccccc|c}
\toprule
uses multi-layer features  & uses domain labels & \textbf{Algorithm} & \textbf{VLCS}  & \textbf{PACS}  & \textbf{OH}   & \textbf{TI}  & \textbf{DN}   & \textbf{Avg}  \\
\midrule
- & - & ERM~\cite{ERM}                       & 76.6           & 83.8           & 67.2          & 47.0         & 44.1          & 63.7          \\
\midrule
\rowcolor{gray!10}\ding{51} & \ding{51} & MLDG~\cite{mldg}                      & 77.2           & 84.9           & 66.8          & 47.7         & 41.2          & 63.6          \\
\rowcolor{gray!10}\ding{51} & \ding{51} & MMD~\cite{mmd}                 & 77.5           & 84.7           & 66.3          & 42.2         & 23.4          & 58.8          \\
\rowcolor{gray!10}\ding{51} & \ding{51} & CORAL~\cite{coral}               & 78.8           & 86.2           & 68.7          & 47.6         & 41.5          & 64.5          \\
\rowcolor{gray!10}\ding{51} & \ding{51} & SagNet~\cite{sagnet}                    & 77.8           & 86.3           & 68.1          & 48.6         & 40.3          & 64.2          \\
\rowcolor{gray!10}\ding{55} & \ding{51} & DANN~\cite{ganin2016domain}                & 78.6           & 83.6           & 65.9          & 46.7         & 38.3          & 62.6          \\
\rowcolor{gray!10}\ding{55} & \ding{51} & Fishr~\cite{rame2022fishr}                    & 77.8           & 85.5           & 67.8          & 47.4         & 41.7          & 64.0          \\
\rowcolor{gray!10}\ding{51} & \ding{55} & MIRO~\cite{miro}                & \textbf{79.0}           & 85.4           & \textbf{70.5}          & 50.4         & 44.3          & 65.9          \\
\rowcolor{gray!10}\ding{55} & \ding{55} & Mixup~\cite{mixup1, mixup2}              & 77.4           & 84.6           & 68.1          & 47.9         & 39.2          & 63.4          \\
\rowcolor{gray!10}\ding{55} & \ding{55} & LatentDR (SA)~\cite{liu2024latentdr}                & 78.7           & 85.8           & 69.0          & 49.9         & 45.1          & 65.7          \\
\rowcolor{gray!10}\ding{55} & \ding{55} & LatentDR (Pool)~\cite{liu2024latentdr}                & 78.0           & 86.3           & 68.4          & 49.5         & 43.9          & 65.2          \\
\midrule
\rowcolor{cyan!10}\ding{55} & \ding{51} & DA-ERM (\cite{Dubey2021AdaptiveMF})             & 78.0           & 84.1  & 67.9          & 47.3         & 43.6          & 64.1          \\
\rowcolor{cyan!10}\ding{55} & \ding{55} & AdaClust (\cite{thomas2021adaptivemethodsaggregateddomain})             & \underline{78.9}           & 87.0  & 67.7          & 48.1         & 43.6         & 64.9          \\
\rowcolor{cyan!10}\ding{55} & \ding{55} & \algoname-DiT (ours)            & 78.5  & 87.1  & 68.4  & 48.2  & 45.8  & 65.6  \\
\rowcolor{cyan!10}\ding{55} & \ding{55} & \algoname-SD-2.1 (ours)          & 77.0  & 86.9  & 68.6  & 51.3  & 45.9  & 65.9 \\
\rowcolor{cyan!10}\ding{55} & \ding{55} & \algoname-BEST (ours)        & 78.5  & \underline{\textbf{87.1}}  & \underline{68.6}  & \underline{\textbf{51.3}}  & \underline{\textbf{45.9}}  & \underline{\textbf{66.3}} \\
\bottomrule
\end{tabular}}
\caption{\footnotesize \textbf{Comparison of {\algoname} with other domain generalization algorithms on \boldmath{$5$} datasets:} 
utilizing the DomainBed test bed. The methods are categorized based on (1) whether they operate across multiple intermediate layers in the network and (2) whether they require explicit ground truth domain labels during training. The highest-performing method that does not rely on either is \underline{underlined}. The overall best-performing method is in \textbf{bold}. Methods in \colorbox{cyan!10}{cyan} corresponds to domain-adaptive classifiers (described in Sec.~\ref{sec:diff_domain_gen}). Among those methods we find {\algoname} achieves the highest performance. {\algoname}-BEST reports the best performance among the two diffusion latent spaces (DiT and SD-2.1) for easy reading.}
\label{tab:domainbed}
\end{table}

\noindent \textbf{Comparison with prior art:} In Table~\ref{tab:domainbed}, we compare {\algoname} with other state-of-the-art domain generalization algorithms\footnote{We compare against algorithms reported in \cite{liu2024latentdr, Dubey2021AdaptiveMF, thomas2021adaptivemethodsaggregateddomain}.} and note that {\algoname}-BEST achieves the highest average performance of $\textbf{66.3\%}$ \underline{without using domain labels at any point}. Compared to all methods, {\algoname}-BEST shows the largest improvements on the PACS, TerraIncognita, and DomainNet datasets. The significant gains on DomainNet, a dataset with over $500,000$ images across $325$ classes and $6$ domains, highlights {\algoname}'s ability to scale to larger datasets. Among the domain-adaptive classifier frameworks (bottom rows), {\algoname}-BEST outperforms DA-ERM~\cite{Dubey2021AdaptiveMF} by $\textbf{+2.2\%}$ and AdaClust~\cite{thomas2021adaptivemethodsaggregateddomain} by $\textbf{+1.4\%}$. Notably, the reported scores for most algorithms are obtained after extensive hyper-parameter searches, whereas {\algoname} achieves these gains with the default setting of DomainBed without using features from multiple layers or ground truth domain labels. Overall, results in Tables~\ref{tab:domain_gen_spaces} and~\ref{tab:domainbed} validate our hypothesis that augmenting a feature space with rich domain-specific information on seen domains results in an overall generalizable feature space for \textit{unseen domains}.

\begin{table}[!t]
\centering
\small
\resizebox{0.45\textwidth}{!}{%
\begin{tabular}{l|cc|cc|cc}
\toprule
\textbf{Dataset} 
& \multicolumn{2}{c|}{\textbf{ERM}} 
& \multicolumn{2}{c|}{\textbf{MIRO}} 
& \multicolumn{2}{c}{\textbf{GUIDE}} \\
\cmidrule(lr){2-3} \cmidrule(lr){4-5} \cmidrule(lr){6-7}
& ERM & ERM++ & + SWAD & + ERM++ & + MIRO + SWAD & + ERM++ \\
\midrule
PACS & 83.8 & 88.0 & 88.4 & 88.8 & 89.0 & \textbf{89.2} \\
TI   & 47.0 & 50.7 & 52.9 & 53.4 & 53.1 & \textbf{53.6} \\
\bottomrule
\end{tabular}
}
\caption{\footnotesize 
\textbf{Comparison using SWAD}~\cite{cha2021swad}\textbf{, MIRO}~\cite{miro}\textbf{, and ERM++}~\cite{teterwak2304erm++} \textbf{on PACS and TerraIncognita (TI).} GUIDE trained with ERM++ further improves performance.
}
\label{tab:plusplus}
\end{table}
\noindent {\bf Effect of enhanced training strategies:} We follow the ERM++~\cite{teterwak2304erm++} implementation from DomainBed~\cite{gulrajani2021in} which improves ERM by better utilization of training data, model parameter selection, and weight-space regularization techniques. From Table~\ref{tab:plusplus}, ERM++ improves over standard ERM by $\textbf{+4.2\%}$ on PACS and $\textbf{+3.7\%}$ on TerraIncognita. Applying the same strategies to {\algoname}, we achieve even greater improvements, with {\algoname} + ERM++ outperforming ERM by $\textbf{+5.4\%}$ on PACS and $\textbf{+6.6\%}$ on TerraIncognita. These results show that {\algoname} could benefit from training optimizations proposed over ERM, such as SWAD~\cite{cha2021swad}, MIRO~\cite{miro}, and ERM++~\cite{teterwak2304erm++}.

\noindent \textbf{Is clustering necessary?}
With a motive to understand the role of clustering of features from $\Psi$ before feature concatenation, we conduct an empirical analysis comparing {\algoname} with and without pseudo-domain clustering. To this end, we directly append the raw features $\bPsi(x)$ to $\Phi(x)$. This results in a moderate gain of $+1.3$ over ERM, whereas clustering improves performance by $\mathbf{+3.3}$ on the PACS dataset. We believe that clustering helps smooth out any noise or sample-specific variations and creates more stable (pseudo) domain representations. Clustering also offers more interpretability to inspect what domain-specific variations are captured in the latent space (Fig.~\ref{fig:pseudo_domain_cluster}).

\begin{figure}[!t]
    \centering
    {\includegraphics[width=0.85\linewidth]{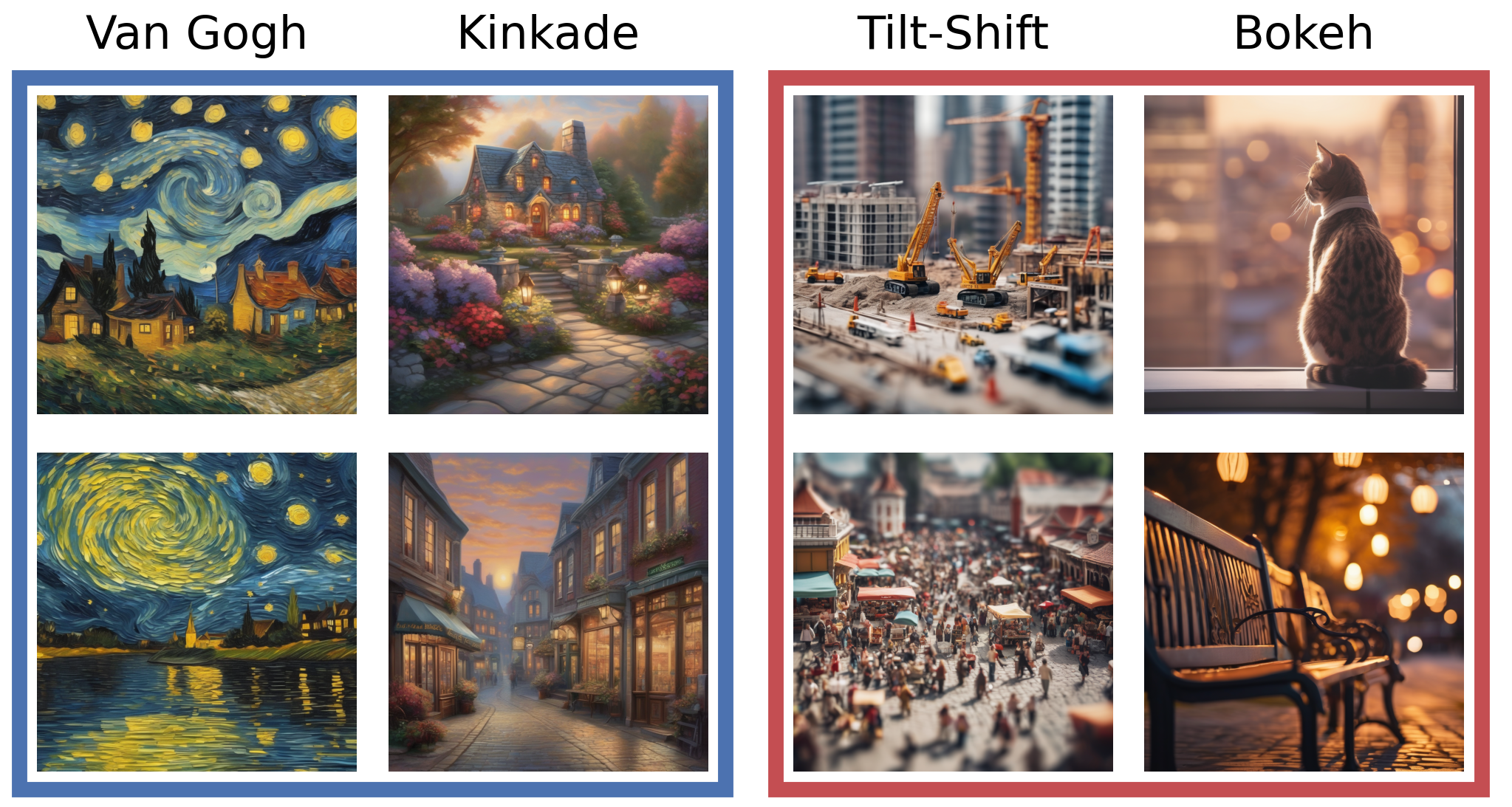}}
    \caption{\footnotesize \textbf{Example images from Synth-Artists and Synth-Photography, generated using Stable Diffusion XL}~\cite{sdxl}. \textcolor{blue}{Synth-Artists} includes artistic styles such as Van Gogh and Kinkade, while the \textcolor{red}{Synth-Photography} captures photography effects like Tilt-Shift and Bokeh.}
    \label{fig:sythn_images}
\end{figure} 
\subsection{Pseudo-domains for Style Discovery}
\label{sec:appl}
Next, we evaluate different pre-training objectives on the task of photographic and artistic style separation. Automatic style identification is valuable for curating and inspecting large-scale datasets, image retrieval, and several such applications. To study this, we first construct two datasets with controlled domain shifts using Stable Diffusion XL~\cite{sdxl} (dataset construction details in suppl. material): (i) \textbf{Synth-Photography} features photographic styles such as macro, tilt-shift, bokeh, symmetry, and zoom blur. Thus, the domain shifts are primarily driven by variations in focus, sharpness, edge details, and depth contrasts. (ii) \textbf{Synth-Artists}, captures styles of Van Gogh, Kinkade, Warhol, Rembrandt, and Dali, making the domain shifts more high-level such as brush stroke patterns and color palettes. We show a few example images in Fig~\ref{fig:sythn_images}. On Synth-Artists, we observe that DiT achieves better domain separation, with a domain NMI score of $0.89$ (Table~\ref{tab:dataset_comparison}). By contrast, on Synth-Photography, SD-2.1 performs better, achieving a domain NMI score of $0.43$ compared to DiT's score of $0.35$ (Table~\ref{tab:dataset_comparison}). This finding aligns with our analysis from Sec.~\ref{sec:pretrain_obj} that 
DiT seem more apt for global variations and SD-2.1 for finer-grained spatially detailed variations.

\section{Discussion and Future Work}
In this work, we introduce {\algoname}, a simple yet effective framework that improves generalization to unseen domains in the absence of domain labels during both train and test times. {\algoname} learns pseudo-domain representations from pre-trained diffusion models and leverages them for domain generalization. Future work includes exploring ways to combine multiple models and build a generalizable latent space that works ``out of the box'' for diverse tasks. 
\clearpage
\newpage
{
    \small
    \bibliographystyle{ieeenat_fullname}
    \bibliography{main}

\begin{thebibliography}{79}
\providecommand{\natexlab}[1]{#1}
\providecommand{\url}[1]{\texttt{#1}}
\expandafter\ifx\csname urlstyle\endcsname\relax
  \providecommand{\doi}[1]{doi: #1}\else
  \providecommand{\doi}{doi: \begingroup \urlstyle{rm}\Url}\fi

\bibitem[Balaji et~al.(2018)Balaji, Sankaranarayanan, and Chellappa]{balaji2018metareg}
Yogesh Balaji, Swami Sankaranarayanan, and Rama Chellappa.
\newblock Metareg: Towards domain generalization using meta-regularization.
\newblock \emph{Advances in Neural Information Processing Systems}, 2018.

\bibitem[Baranchuk et~al.(2021)Baranchuk, Rubachev, Voynov, Khrulkov, and Babenko]{label_seg}
Dmitry Baranchuk, Ivan Rubachev, Andrey Voynov, Valentin Khrulkov, and Artem Babenko.
\newblock Label-efficient semantic segmentation with diffusion models.
\newblock \emph{arXiv preprint arXiv:2112.03126}, 2021.

\bibitem[Beery et~al.(2018)Beery, Van~Horn, and Perona]{TerraInc}
Sara Beery, Grant Van~Horn, and Pietro Perona.
\newblock Recognition in terra incognita.
\newblock In \emph{Proceedings of the European conference on computer vision (ECCV)}, 2018.

\bibitem[Ben-David et~al.(2010)Ben-David, Blitzer, Crammer, Kulesza, Pereira, and Vaughan]{ben2010theory}
Shai Ben-David, John Blitzer, Koby Crammer, Alex Kulesza, Fernando Pereira, and Jennifer~Wortman Vaughan.
\newblock A theory of learning from different domains.
\newblock \emph{Machine learning}, 2010.

\bibitem[Blanchard et~al.(2011)Blanchard, Lee, and Scott]{Blanchard2011GeneralizingFS}
Gilles Blanchard, Gyemin Lee, and Clayton Scott.
\newblock Generalizing from several related classification tasks to a new unlabeled sample.
\newblock \emph{Advances in Neural Information Processing Systems}, 2011.

\bibitem[Bui et~al.(2021)Bui, Tran, Tran, and Phung]{bui2021exploiting}
Manh-Ha Bui, Toan Tran, Anh Tran, and Dinh Phung.
\newblock Exploiting domain-specific features to enhance domain generalization.
\newblock \emph{Advances in Neural Information Processing Systems}, 2021.

\bibitem[Caron et~al.(2018)Caron, Bojanowski, Joulin, and Douze]{deepcluster}
Mathilde Caron, Piotr Bojanowski, Armand Joulin, and Matthijs Douze.
\newblock Deep clustering for unsupervised learning of visual features.
\newblock In \emph{Proceedings of the European conference on computer vision (ECCV)}, 2018.

\bibitem[Caron et~al.(2020)Caron, Misra, Mairal, Goyal, Bojanowski, and Joulin]{swav}
Mathilde Caron, Ishan Misra, Julien Mairal, Priya Goyal, Piotr Bojanowski, and Armand Joulin.
\newblock Unsupervised learning of visual features by contrasting cluster assignments.
\newblock \emph{Advances in Neural Information Processing Systems}, 2020.

\bibitem[Cha et~al.(2021)Cha, Chun, Lee, Cho, Park, Lee, and Park]{cha2021swad}
Junbum Cha, Sanghyuk Chun, Kyungjae Lee, Han-Cheol Cho, Seunghyun Park, Yunsung Lee, and Sungrae Park.
\newblock Swad: Domain generalization by seeking flat minima.
\newblock \emph{Advances in Neural Information Processing Systems}, 2021.

\bibitem[Cha et~al.(2022)Cha, Lee, Park, and Chun]{miro}
Junbum Cha, Kyungjae Lee, Sungrae Park, and Sanghyuk Chun.
\newblock Domain generalization by mutual-information regularization with pre-trained models.
\newblock In \emph{Proceedings of the European conference on computer vision (ECCV)}, 2022.

\bibitem[Chen et~al.(2023)Chen, Sun, Song, and Luo]{diffusion_detect}
Shoufa Chen, Peize Sun, Yibing Song, and Ping Luo.
\newblock Diffusiondet: Diffusion model for object detection.
\newblock In \emph{Proceedings of the IEEE/CVF international conference on computer vision}, pages 19830--19843, 2023.

\bibitem[Chen et~al.(2020)Chen, Kornblith, Norouzi, and Hinton]{simclr}
Ting Chen, Simon Kornblith, Mohammad Norouzi, and Geoffrey Hinton.
\newblock A simple framework for contrastive learning of visual representations.
\newblock In \emph{International conference on machine learning}, 2020.

\bibitem[Deng et~al.(2009)Deng, Dong, Socher, Li, Li, and Fei-Fei]{deng2009imagenet}
Jia Deng, Wei Dong, Richard Socher, Li-Jia Li, Kai Li, and Li Fei-Fei.
\newblock Imagenet: A large-scale hierarchical image database.
\newblock In \emph{Proceedings of the IEEE/CVF conference on computer vision and pattern recognition}, 2009.

\bibitem[Dubey et~al.(2021)Dubey, Ramanathan, Pentland, and Mahajan]{Dubey2021AdaptiveMF}
Abhimanyu Dubey, Vignesh Ramanathan, Alex Pentland, and Dhruv Mahajan.
\newblock Adaptive methods for real-world domain generalization.
\newblock In \emph{Proceedings of the IEEE/CVF conference on computer vision and pattern recognition}, 2021.

\bibitem[Fang et~al.(2013)Fang, Xu, and Rockmore]{VLCS}
Chen Fang, Ye Xu, and Daniel~N Rockmore.
\newblock Unbiased metric learning: On the utilization of multiple datasets and web images for softening bias.
\newblock In \emph{Proceedings of the IEEE/CVF international conference on computer vision}, 2013.

\bibitem[Ganin et~al.(2016)Ganin, Ustinova, Ajakan, Germain, Larochelle, Laviolette, March, and Lempitsky]{ganin2016domain}
Yaroslav Ganin, Evgeniya Ustinova, Hana Ajakan, Pascal Germain, Hugo Larochelle, Fran{\c{c}}ois Laviolette, Mario March, and Victor Lempitsky.
\newblock Domain-adversarial training of neural networks.
\newblock \emph{Journal of machine learning research}, 2016.

\bibitem[Gulrajani and Lopez-Paz(2021)]{gulrajani2021in}
Ishaan Gulrajani and David Lopez-Paz.
\newblock In search of lost domain generalization.
\newblock In \emph{International Conference on Learning Representations}, 2021.

\bibitem[He et~al.(2016)He, Zhang, Ren, and Sun]{rn50}
Kaiming He, Xiangyu Zhang, Shaoqing Ren, and Jian Sun.
\newblock Deep residual learning for image recognition.
\newblock \emph{Proceedings of the IEEE/CVF conference on computer vision and pattern recognition}, 2016.

\bibitem[He et~al.(2020)He, Fan, Wu, Xie, and Girshick]{moco}
Kaiming He, Haoqi Fan, Yuxin Wu, Saining Xie, and Ross Girshick.
\newblock Momentum contrast for unsupervised visual representation learning.
\newblock In \emph{Proceedings of the IEEE/CVF conference on computer vision and pattern recognition}, 2020.

\bibitem[He et~al.(2022)He, Chen, Xie, Li, Doll{\'a}r, and Girshick]{MAE}
Kaiming He, Xinlei Chen, Saining Xie, Yanghao Li, Piotr Doll{\'a}r, and Ross Girshick.
\newblock Masked autoencoders are scalable vision learners.
\newblock In \emph{Proceedings of the IEEE/CVF conference on computer vision and pattern recognition}, 2022.

\bibitem[Hemati et~al.(2022)Hemati, Beitollahi, Estiri, Al~Omari, Lamghari, Khalil, Chen, and Zhang]{hematibeyond}
Sobhan Hemati, Mahdi Beitollahi, Amir~Hossein Estiri, Bassel Al~Omari, Soufiane Lamghari, Yasser~H Khalil, Xi Chen, and Guojun Zhang.
\newblock Beyond loss functions: Exploring data-centric approaches with diffusion model for domain generalization.
\newblock \emph{Transactions on Machine Learning Research}, 2022.

\bibitem[Hemati et~al.(2023)Hemati, Beitollahi, Estiri, Omari, Chen, and Zhang]{hemati2023cross}
Sobhan Hemati, Mahdi Beitollahi, Amir~Hossein Estiri, Bassel~Al Omari, Xi Chen, and Guojun Zhang.
\newblock Cross domain generative augmentation: Domain generalization with latent diffusion models.
\newblock \emph{arXiv preprint arXiv:2312.05387}, 2023.

\bibitem[Hendrycks et~al.(2019)Hendrycks, Mu, Cubuk, Zoph, Gilmer, and Lakshminarayanan]{hendrycks2019augmix}
Dan Hendrycks, Norman Mu, Ekin~D Cubuk, Barret Zoph, Justin Gilmer, and Balaji Lakshminarayanan.
\newblock Augmix: A simple data processing method to improve robustness and uncertainty.
\newblock \emph{arXiv preprint arXiv:1912.02781}, 2019.

\bibitem[Ho et~al.(2020)Ho, Jain, and Abbeel]{ddpm}
Jonathan Ho, Ajay Jain, and Pieter Abbeel.
\newblock Denoising diffusion probabilistic models.
\newblock \emph{Advances in Neural Information Processing Systems}, 2020.

\bibitem[Hong et~al.(2021)Hong, Choi, and Kim]{hong2021stylemix}
Minui Hong, Jinwoo Choi, and Gunhee Kim.
\newblock Stylemix: Separating content and style for enhanced data augmentation.
\newblock In \emph{Proceedings of the IEEE/CVF conference on computer vision and pattern recognition}, 2021.

\bibitem[Huang et~al.(2025)Huang, Chen, Liu, Zhang, Dai, Xiong, and Tian]{huang2025domainfusion}
Yuyang Huang, Yabo Chen, Yuchen Liu, Xiaopeng Zhang, Wenrui Dai, Hongkai Xiong, and Qi Tian.
\newblock Domainfusion: Generalizing to unseen domains with latent diffusion models.
\newblock In \emph{Proceedings of the European conference on computer vision (ECCV)}, 2025.

\bibitem[Huang et~al.(2020)Huang, Wang, Xing, and Huang]{huang2020self}
Zeyi Huang, Haohan Wang, Eric~P Xing, and Dong Huang.
\newblock Self-challenging improves cross-domain generalization.
\newblock In \emph{Proceedings of the European conference on computer vision (ECCV)}, 2020.

\bibitem[Jia et~al.(2021)Jia, Yang, Xia, Chen, Parekh, Pham, Le, Sung, Li, and Duerig]{ALIGN}
Chao Jia, Yinfei Yang, Ye Xia, Yi-Ting Chen, Zarana Parekh, Hieu Pham, Quoc Le, Yun-Hsuan Sung, Zhen Li, and Tom Duerig.
\newblock Scaling up visual and vision-language representation learning with noisy text supervision.
\newblock In \emph{International conference on machine learning}, 2021.

\bibitem[Jiang et~al.(2023)Jiang, Liu, Liu, Zhao, Zhang, Gao, Zhang, Li, and Xiong]{jiang2023clip}
Dongsheng Jiang, Yuchen Liu, Songlin Liu, Jin'e Zhao, Hao Zhang, Zhen Gao, Xiaopeng Zhang, Jin Li, and Hongkai Xiong.
\newblock From clip to dino: Visual encoders shout in multi-modal large language models.
\newblock \emph{arXiv preprint arXiv:2310.08825}, 2023.

\bibitem[Kim et~al.(2024)Kim, Thomas, and Ghadiyaram]{revelio}
Dahye Kim, Xavier Thomas, and Deepti Ghadiyaram.
\newblock Revelio: Interpreting and leveraging semantic information in diffusion models.
\newblock \emph{arXiv preprint arXiv:2411.16725}, 2024.

\bibitem[Kingma et~al.(2013)Kingma, Welling, et~al.]{kingma2013auto}
Diederik~P Kingma, Max Welling, et~al.
\newblock Auto-encoding variational bayes, 2013.

\bibitem[Li et~al.(2023)Li, Prabhudesai, Duggal, Brown, and Pathak]{diffusion_classifier_23}
Alexander~C Li, Mihir Prabhudesai, Shivam Duggal, Ellis Brown, and Deepak Pathak.
\newblock Your diffusion model is secretly a zero-shot classifier.
\newblock In \emph{Proceedings of the IEEE/CVF International Conference on Computer Vision}, pages 2206--2217, 2023.

\bibitem[Li et~al.(2017)Li, Yang, Song, and Hospedales]{PACS}
Da Li, Yongxin Yang, Yi-Zhe Song, and Timothy~M Hospedales.
\newblock Deeper, broader and artier domain generalization.
\newblock In \emph{Proceedings of the IEEE international conference on computer vision}, 2017.

\bibitem[Li et~al.(2018{\natexlab{a}})Li, Yang, Song, and Hospedales]{mldg}
Da Li, Yongxin Yang, Yi-Zhe Song, and Timothy Hospedales.
\newblock Learning to generalize: Meta-learning for domain generalization.
\newblock In \emph{Proceedings of the AAAI conference on artificial intelligence}, 2018{\natexlab{a}}.

\bibitem[Li et~al.(2018{\natexlab{b}})Li, Pan, Wang, and Kot]{mmd}
Haoliang Li, Sinno~Jialin Pan, Shiqi Wang, and Alex~C Kot.
\newblock Domain generalization with adversarial feature learning.
\newblock In \emph{Proceedings of the IEEE/CVF conference on computer vision and pattern recognition}, 2018{\natexlab{b}}.

\bibitem[Li et~al.(2024)Li, Yu, Du, Zhu, and Shen]{source_free_da}
Jingjing Li, Zhiqi Yu, Zhekai Du, Lei Zhu, and Heng~Tao Shen.
\newblock A comprehensive survey on source-free domain adaptation.
\newblock \emph{IEEE Transactions on Pattern Analysis and Machine Intelligence}, 2024.

\bibitem[Li et~al.(2021)Li, Li, Li, Gong, Fu, and Hospedales]{li2021simple}
Pan Li, Da Li, Wei Li, Shaogang Gong, Yanwei Fu, and Timothy~M Hospedales.
\newblock A simple feature augmentation for domain generalization.
\newblock In \emph{Proceedings of the IEEE international conference on computer vision}, 2021.

\bibitem[Liang et~al.(2022)Liang, Li, and Marculescu]{liang2022supmae}
Feng Liang, Yangguang Li, and Diana Marculescu.
\newblock Supmae: Supervised masked autoencoders are efficient vision learners.
\newblock \emph{arXiv preprint arXiv:2205.14540}, 2022.

\bibitem[Liu et~al.(2023)Liu, Li, Wu, and Lee]{llava}
Haotian Liu, Chunyuan Li, Qingyang Wu, and Yong~Jae Lee.
\newblock Visual instruction tuning.
\newblock \emph{Advances in neural information processing systems}, 36:\penalty0 34892--34916, 2023.

\bibitem[Liu et~al.(2024)Liu, Khose, Xiao, Sathidevi, Ramnath, Kira, and Dyer]{liu2024latentdr}
Ran Liu, Sahil Khose, Jingyun Xiao, Lakshmi Sathidevi, Keerthan Ramnath, Zsolt Kira, and Eva~L Dyer.
\newblock Latentdr: Improving model generalization through sample-aware latent degradation and restoration.
\newblock In \emph{Proceedings of the IEEE Winter Conference on Applications of Computer Vision}, 2024.

\bibitem[Luo et~al.(2024)Luo, Dunlap, Park, Holynski, and Darrell]{diff_hyper}
Grace Luo, Lisa Dunlap, Dong~Huk Park, Aleksander Holynski, and Trevor Darrell.
\newblock Diffusion hyperfeatures: Searching through time and space for semantic correspondence.
\newblock \emph{Advances in Neural Information Processing Systems}, 2024.

\bibitem[Mahajan et~al.(2018)Mahajan, Girshick, Ramanathan, He, Paluri, Li, Bharambe, and Van Der~Maaten]{mahajan2018exploring}
Dhruv Mahajan, Ross Girshick, Vignesh Ramanathan, Kaiming He, Manohar Paluri, Yixuan Li, Ashwin Bharambe, and Laurens Van Der~Maaten.
\newblock Exploring the limits of weakly supervised pretraining.
\newblock In \emph{Proceedings of the European conference on computer vision (ECCV)}, 2018.

\bibitem[Matsuura and Harada(2020)]{matsuura2020domain}
Toshihiko Matsuura and Tatsuya Harada.
\newblock Domain generalization using a mixture of multiple latent domains.
\newblock In \emph{Proceedings of the AAAI conference on artificial intelligence}, 2020.

\bibitem[Muandet et~al.(2017)Muandet, Fukumizu, Sriperumbudur, Sch{\"o}lkopf, et~al.]{muandet2017kernel}
Krikamol Muandet, Kenji Fukumizu, Bharath Sriperumbudur, Bernhard Sch{\"o}lkopf, et~al.
\newblock Kernel mean embedding of distributions: A review and beyond.
\newblock \emph{Foundations and Trends{\textregistered} in Machine Learning}, 2017.

\bibitem[Nam et~al.(2021{\natexlab{a}})Nam, Lee, Park, Yoon, and Yoo]{nam2021reducing}
Hyeonseob Nam, HyunJae Lee, Jongchan Park, Wonjun Yoon, and Donggeun Yoo.
\newblock Reducing domain gap by reducing style bias.
\newblock In \emph{Proceedings of the IEEE/CVF conference on computer vision and pattern recognition}, 2021{\natexlab{a}}.

\bibitem[Nam et~al.(2021{\natexlab{b}})Nam, Lee, Park, Yoon, and Yoo]{sagnet}
Hyeonseob Nam, HyunJae Lee, Jongchan Park, Wonjun Yoon, and Donggeun Yoo.
\newblock Reducing domain gap by reducing style bias.
\newblock In \emph{Proceedings of the IEEE/CVF conference on computer vision and pattern recognition}, 2021{\natexlab{b}}.

\bibitem[Oquab et~al.(2023)Oquab, Darcet, Moutakanni, Vo, Szafraniec, Khalidov, Fernandez, Haziza, Massa, El-Nouby, et~al.]{Oquab2023DINOv2LR}
Maxime Oquab, Timoth{\'e}e Darcet, Th{\'e}o Moutakanni, Huy Vo, Marc Szafraniec, Vasil Khalidov, Pierre Fernandez, Daniel Haziza, Francisco Massa, Alaaeldin El-Nouby, et~al.
\newblock Dinov2: Learning robust visual features without supervision.
\newblock \emph{arXiv preprint arXiv:2304.07193}, 2023.

\bibitem[Park et~al.(2023)Park, Kwon, Choi, Jo, and Uh]{park2023understanding}
Yong-Hyun Park, Mingi Kwon, Jaewoong Choi, Junghyo Jo, and Youngjung Uh.
\newblock Understanding the latent space of diffusion models through the lens of riemannian geometry.
\newblock \emph{Advances in Neural Information Processing Systems}, 2023.

\bibitem[Peebles and Xie(2023)]{dit}
William Peebles and Saining Xie.
\newblock Scalable diffusion models with transformers.
\newblock In \emph{Proceedings of the IEEE/CVF international conference on computer vision}, pages 4195--4205, 2023.

\bibitem[Peng et~al.(2019)Peng, Bai, Xia, Huang, Saenko, and Wang]{domainnet}
Xingchao Peng, Qinxun Bai, Xide Xia, Zijun Huang, Kate Saenko, and Bo Wang.
\newblock Moment matching for multi-source domain adaptation.
\newblock In \emph{Proceedings of the IEEE international conference on computer vision}, 2019.

\bibitem[Podell et~al.(2023)Podell, English, Lacey, Blattmann, Dockhorn, M{\"u}ller, Penna, and Rombach]{sdxl}
Dustin Podell, Zion English, Kyle Lacey, Andreas Blattmann, Tim Dockhorn, Jonas M{\"u}ller, Joe Penna, and Robin Rombach.
\newblock Sdxl: Improving latent diffusion models for high-resolution image synthesis.
\newblock \emph{arXiv preprint arXiv:2307.01952}, 2023.

\bibitem[Qian et~al.(2024)Qian, Cai, Pan, Li, Yao, Sun, and Mei]{qian2024boosting}
Yurui Qian, Qi Cai, Yingwei Pan, Yehao Li, Ting Yao, Qibin Sun, and Tao Mei.
\newblock Boosting diffusion models with moving average sampling in frequency domain.
\newblock In \emph{Proceedings of the IEEE/CVF conference on computer vision and pattern recognition}, 2024.

\bibitem[Radford et~al.(2021)Radford, Kim, Hallacy, Ramesh, Goh, Agarwal, Sastry, Askell, Mishkin, Clark, et~al.]{pmlr-v139-radford21a}
Alec Radford, Jong~Wook Kim, Chris Hallacy, Aditya Ramesh, Gabriel Goh, Sandhini Agarwal, Girish Sastry, Amanda Askell, Pamela Mishkin, Jack Clark, et~al.
\newblock Learning transferable visual models from natural language supervision.
\newblock In \emph{International conference on machine learning}, pages 8748--8763. PmLR, 2021.

\bibitem[Rame et~al.(2022)Rame, Dancette, and Cord]{rame2022fishr}
Alexandre Rame, Corentin Dancette, and Matthieu Cord.
\newblock Fishr: Invariant gradient variances for out-of-distribution generalization.
\newblock In \emph{International conference on machine learning}, 2022.

\bibitem[Recht et~al.(2019)Recht, Roelofs, Schmidt, and Shankar]{recht2019imagenet}
Benjamin Recht, Rebecca Roelofs, Ludwig Schmidt, and Vaishaal Shankar.
\newblock Do imagenet classifiers generalize to imagenet?
\newblock In \emph{International conference on machine learning}, 2019.

\bibitem[Rombach et~al.(2022)Rombach, Blattmann, Lorenz, Esser, and Ommer]{stablediffusion}
Robin Rombach, Andreas Blattmann, Dominik Lorenz, Patrick Esser, and Bj{\"o}rn Ommer.
\newblock High-resolution image synthesis with latent diffusion models.
\newblock In \emph{Proceedings of the IEEE/CVF conference on computer vision and pattern recognition}, 2022.

\bibitem[Ronneberger et~al.(2015)Ronneberger, Fischer, and Brox]{unet}
Olaf Ronneberger, Philipp Fischer, and Thomas Brox.
\newblock U-net: Convolutional networks for biomedical image segmentation.
\newblock In \emph{Medical image computing and computer-assisted intervention}, 2015.

\bibitem[Sohl-Dickstein et~al.(2015)Sohl-Dickstein, Weiss, Maheswaranathan, and Ganguli]{sohl2015deep}
Jascha Sohl-Dickstein, Eric Weiss, Niru Maheswaranathan, and Surya Ganguli.
\newblock Deep unsupervised learning using nonequilibrium thermodynamics.
\newblock In \emph{International conference on machine learning}. PMLR, 2015.

\bibitem[Somavarapu et~al.(2020)Somavarapu, Ma, and Kira]{somavarapu2020frustratingly}
Nathan Somavarapu, Chih-Yao Ma, and Zsolt Kira.
\newblock Frustratingly simple domain generalization via image stylization.
\newblock \emph{arXiv preprint arXiv:2006.11207}, 2020.

\bibitem[Sun and Saenko(2016)]{coral}
Baochen Sun and Kate Saenko.
\newblock Deep coral: Correlation alignment for deep domain adaptation.
\newblock In \emph{Proceedings of the European conference on computer vision (ECCV)}, 2016.

\bibitem[Taori et~al.(2020)Taori, Dave, Shankar, Carlini, Recht, and Schmidt]{taori}
Rohan Taori, Achal Dave, Vaishaal Shankar, Nicholas Carlini, Benjamin Recht, and Ludwig Schmidt.
\newblock Measuring robustness to natural distribution shifts in image classification.
\newblock \emph{Advances in Neural Information Processing Systems}, 2020.

\bibitem[Teterwak et~al.(2023)Teterwak, Saito, Tsiligkaridis, Saenko, and Plummer]{teterwak2304erm++}
Piotr Teterwak, Kuniaki Saito, Theodoros Tsiligkaridis, Kate Saenko, and Bryan~A Plummer.
\newblock Erm++: An improved baseline for domain generalization.
\newblock \emph{arXiv preprint arXiv.2304.01973}, 2023.

\bibitem[Teterwak et~al.(2024)Teterwak, Saito, Tsiligkaridis, Plummer, and Saenko]{teterwak2024large}
Piotr Teterwak, Kuniaki Saito, Theodoros Tsiligkaridis, Bryan~A Plummer, and Kate Saenko.
\newblock Is large-scale pretraining the secret to good domain generalization?
\newblock \emph{arXiv preprint arXiv:2412.02856}, 2024.

\bibitem[Thomas et~al.(2021)Thomas, Mahajan, Pentland, and Dubey]{thomas2021adaptivemethodsaggregateddomain}
Xavier Thomas, Dhruv Mahajan, Alex Pentland, and Abhimanyu Dubey.
\newblock Adaptive methods for aggregated domain generalization.
\newblock \emph{arXiv preprint arXiv:2112.04766}, 2021.

\bibitem[Tong et~al.(2024)Tong, Liu, Zhai, Ma, LeCun, and Xie]{eyeshut}
Shengbang Tong, Zhuang Liu, Yuexiang Zhai, Yi Ma, Yann LeCun, and Saining Xie.
\newblock Eyes wide shut? exploring the visual shortcomings of multimodal llms.
\newblock In \emph{Proceedings of the IEEE/CVF conference on computer vision and pattern recognition}, 2024.

\bibitem[Tumanyan et~al.(2023)Tumanyan, Geyer, Bagon, and Dekel]{plugandplay}
N. Tumanyan, M. Geyer, S. Bagon, and T. Dekel.
\newblock Plug-and-play diffusion features for text-driven image-to-image translation.
\newblock In \emph{Proceedings of the IEEE/CVF conference on computer vision and pattern recognition}, 2023.

\bibitem[Vapnik(1999)]{ERM}
Vladimir~N Vapnik.
\newblock An overview of statistical learning theory.
\newblock \emph{IEEE transactions on neural networks}, 1999.

\bibitem[Venkateswara et~al.(2017)Venkateswara, Eusebio, Chakraborty, and Panchanathan]{OfficeHome}
Hemanth Venkateswara, Jose Eusebio, Shayok Chakraborty, and Sethuraman Panchanathan.
\newblock Deep hashing network for unsupervised domain adaptation.
\newblock In \emph{Proceedings of the IEEE/CVF conference on computer vision and pattern recognition}, 2017.

\bibitem[Voynov et~al.(2023)Voynov, Chu, Cohen-Or, and Aberman]{voynov2023p+}
Andrey Voynov, Qinghao Chu, Daniel Cohen-Or, and Kfir Aberman.
\newblock p+: Extended textual conditioning in text-to-image generation.
\newblock \emph{arXiv preprint arXiv:2303.09522}, 2023.

\bibitem[Wang et~al.(2024)Wang, Sun, Zhang, Tang, Liu, and Wang]{diva}
Wenxuan Wang, Quan Sun, Fan Zhang, Yepeng Tang, Jing Liu, and Xinlong Wang.
\newblock Diffusion feedback helps clip see better.
\newblock \emph{arXiv preprint arXiv:2407.20171}, 2024.

\bibitem[Wu et~al.(2023)Wu, Zhao, Chen, Gu, Zhao, He, Zhou, Shou, and Shen]{wu2023datasetdm}
Weijia Wu, Yuzhong Zhao, Hao Chen, Yuchao Gu, Rui Zhao, Yefei He, Hong Zhou, Mike~Zheng Shou, and Chunhua Shen.
\newblock Datasetdm: Synthesizing data with perception annotations using diffusion models.
\newblock In \emph{Advances in Neural Information Processing Systems}, 2023.

\bibitem[Xu et~al.(2023)Xu, Liu, Vahdat, Byeon, Wang, and De~Mello]{diffusion_segmentation}
Jiarui Xu, Sifei Liu, Arash Vahdat, Wonmin Byeon, Xiaolong Wang, and Shalini De~Mello.
\newblock Open-vocabulary panoptic segmentation with text-to-image diffusion models.
\newblock In \emph{Proceedings of the IEEE/CVF conference on computer vision and pattern recognition}, 2023.

\bibitem[Xu et~al.(2020)Xu, Zhang, Ni, Li, Wang, Tian, and Zhang]{mixup1}
Minghao Xu, Jian Zhang, Bingbing Ni, Teng Li, Chengjie Wang, Qi Tian, and Wenjun Zhang.
\newblock Adversarial domain adaptation with domain mixup.
\newblock In \emph{Proceedings of the AAAI conference on artificial intelligence}, 2020.

\bibitem[Yan et~al.(2020{\natexlab{a}})Yan, Song, Li, Zou, and Ren]{mixup2}
Shen Yan, Huan Song, Nanxiang Li, Lincan Zou, and Liu Ren.
\newblock Improve unsupervised domain adaptation with mixup training.
\newblock \emph{arXiv preprint arXiv:2001.00677}, 2020{\natexlab{a}}.

\bibitem[Yan et~al.(2020{\natexlab{b}})Yan, Misra, Gupta, Ghadiyaram, and Mahajan]{clusterfit}
Xueting Yan, Ishan Misra, Abhinav Gupta, Deepti Ghadiyaram, and Dhruv Mahajan.
\newblock Clusterfit: Improving generalization of visual representations.
\newblock In \emph{Proceedings of the IEEE/CVF conference on computer vision and pattern recognition}, 2020{\natexlab{b}}.

\bibitem[Yu et~al.(2023)Yu, Liu, Yang, and Wang]{yu2023distribution}
Runpeng Yu, Songhua Liu, Xingyi Yang, and Xinchao Wang.
\newblock Distribution shift inversion for out-of-distribution prediction.
\newblock In \emph{Proceedings of the IEEE/CVF conference on computer vision and pattern recognition}, 2023.

\bibitem[Zhang et~al.(2018)Zhang, Wang, Cai, and Song]{Zhang2018MCA}
Yun Zhang, Nianbin Wang, Shaobin Cai, and Lei Song.
\newblock Unsupervised domain adaptation by mapped correlation alignment.
\newblock \emph{IEEE Access}, 2018.

\bibitem[Zhao et~al.(2023)Zhao, Rao, Liu, Liu, Zhou, and Lu]{zhao2023unleashing}
Wenliang Zhao, Yongming Rao, Zuyan Liu, Benlin Liu, Jie Zhou, and Jiwen Lu.
\newblock Unleashing text-to-image diffusion models for visual perception.
\newblock In \emph{Proceedings of the IEEE international conference on computer vision}, 2023.

\bibitem[Zhenda et~al.(2022)Zhenda, Zigang, Jingcheng, Zheng, Han, and Yue]{zhenda2022revealing}
Xie Zhenda, Geng Zigang, Hu Jingcheng, Zhang Zheng, Hu Han, and Cao Yue.
\newblock Revealing the dark secrets of masked image modeling.
\newblock \emph{arXiv preprint arXiv:2205.13543}, 2022.

\end{thebibliography}
}

\clearpage
\newpage
\onecolumn  
\appendix   
\begin{center}
    {\bf \Large Supplementary Material: \\ What's in a Latent? Leveraging Diffusion Latent Space for Domain Generalization}
\end{center}
\vspace{8pt}



\section{Transformation Function}
\begin{table}[!h]
\centering
\resizebox{0.30\textwidth}{!}{%
\begin{tabular}{lc}
\toprule
\textbf{Transformation ($\mathcal{T}$)} & \textbf{Acc} \\
\midrule
ERM & 83.8 \\
\midrule
Direct Concatenation (No Transformation)      & 84.3                  \\
Cluster-Based Replacement      &     84.6              \\
Linear Regression     &     85.7              \\
RBF Kernel Ridge Regression   &  \textbf{87.1}                 \\
\bottomrule
\end{tabular}}
\caption{\footnotesize \textbf{Effect of $\mathcal{T}$ on Test Accuracy for PACS, using \algoname-DiT.} We find that the RBF step (Sec~\ref{sec:impl}) aids in classification performance on unseen domains.}
\label{tab:concat_ablations}
\end{table}

\noindent {\bf Effect of the choice of $\mathcal{T}$:} 

As noted in Sec.~\ref{sec:diff_domain_gen}, we apply a transformation function $\mathcal{T}:\bPsi \mapsto \bPhi$ to bring the latent manifold of $\Psi$ closer to $\Phi$ and mitigate feature domain drift. To understand the role of $\mathcal{T}$, we explore the following alternatives to it: 
\begin{itemize}
    \item \textbf{(a) Direct concatenation}, i.e., appending pseudo-domain representations (from $\bPsi$) to the features  (from $\bPhi$) without any transformation. While this introduces domain-specific information, lack of alignment between the two feature spaces led to a minimal improvement of $+0.5\%$ over ERM.
    \item \textbf{(b) Cluster-based replacement}, where pseudo-domains identified in the $\bPsi$ space are used to compute cluster centroids using features from $\bPhi$ space, i.e. cluster samples are averaged in $\bPhi$ space. This provides a slightly better alignment yielding an accuracy gain of $+0.8\%$ over the baseline. 
    \item \textbf{(c) Linear regression}, where a linear mapping is learned between the pseudo-domain centroids and the centroids obtained in \textbf{(b)}. This helps in bridging differences between $\bPsi$ and $\bPhi$ better, leading to a larger improvement of $+1.4\%$.
    \item \textbf{(d) RBF kernel ridge regression}, where the linear regressor in \textbf{(c)} is replaced with an RBF kernel  (Sec~\ref{sec:impl}). We note that this achieves the highest accuracy gains of $+3.3\%$, highlighting its effectiveness of bridging feature domain drift while incorporating pseudo-domain information into the classifier.
\end{itemize}

These results underscore the necessity of a well-chosen transformation to fully leverage the pseudo-domain information.

\section{Domain Predictability}
\begin{table}[!h]
\centering
\resizebox{0.55\textwidth}{!}{%
\begin{tabular}{lcccccc}
\toprule
\textbf{Dataset} & \textbf{DiT} & \textbf{SD-2.1} & \textbf{MAE} & \textbf{CLIP} & \textbf{DINOv2} & \textbf{RN50} \\
\midrule
PACS             & 98.89            & \textbf{98.95}           & 98.69            & 98.29            & 98.89             & 97.85             \\
VLCS             & \textbf{96.08}            & 92.72           & 94.03            & 83.87             & 81.86             & 88.48             \\
TerraInc   & \textbf{99.97}           & 99.94           & 99.91           & 99.83            & 99.87             & 99.79            \\
OfficeHome   & \textbf{89.16}            & 86.43           & 82.55           & 83.41           & 78.28             & 77.52           \\
DomainNet   & 88.55            & \textbf{89.58}           & 87.50           & 87.61           & 87.24             & 87.21           \\
Synth-Artists & \textbf{100}           & 99.00          & 97.00            & 92.00             & 90.00             & 97.00           \\
Synth-Photography & 83.33           & \textbf{87.50}           & 86.67            & 73.33             & 78.33             & 77.50            \\
\bottomrule
\end{tabular}}
\caption{\footnotesize \textbf{Comparison of Domain Predictability Scores Across Datasets.} Diffusion models consistently outperform other models in domain predictability scores, highlighting the effectiveness of encoding domain-specific information in their latent space.}
\label{tab:domain_pred_comparison}
\end{table}
\noindent\textbf{Domain Predictability:} To complement NMI, we evaluate domain predictability and predict domain labels from latent feature representations. Specifically, we use a single-layer MLP classifier, trained on an 80-20 train-test split. We report the mean test accuracy over 3 such random splits. While NMI measures alignment and variance across samples belonging to a domain, domain predictability directly assesses a latent representation's ability to learn to classify domain information.
We observe in Table.~\ref{tab:domain_pred_comparison} that diffusion models attain the highest domain predictability scores, highlighting their effectiveness in encoding domain-specific information.
\FloatBarrier

\clearpage
\section{Label Noise and Domain Inconsistencies}
\begin{figure}[!h]
    \centering
    {\includegraphics[width=0.85\linewidth]{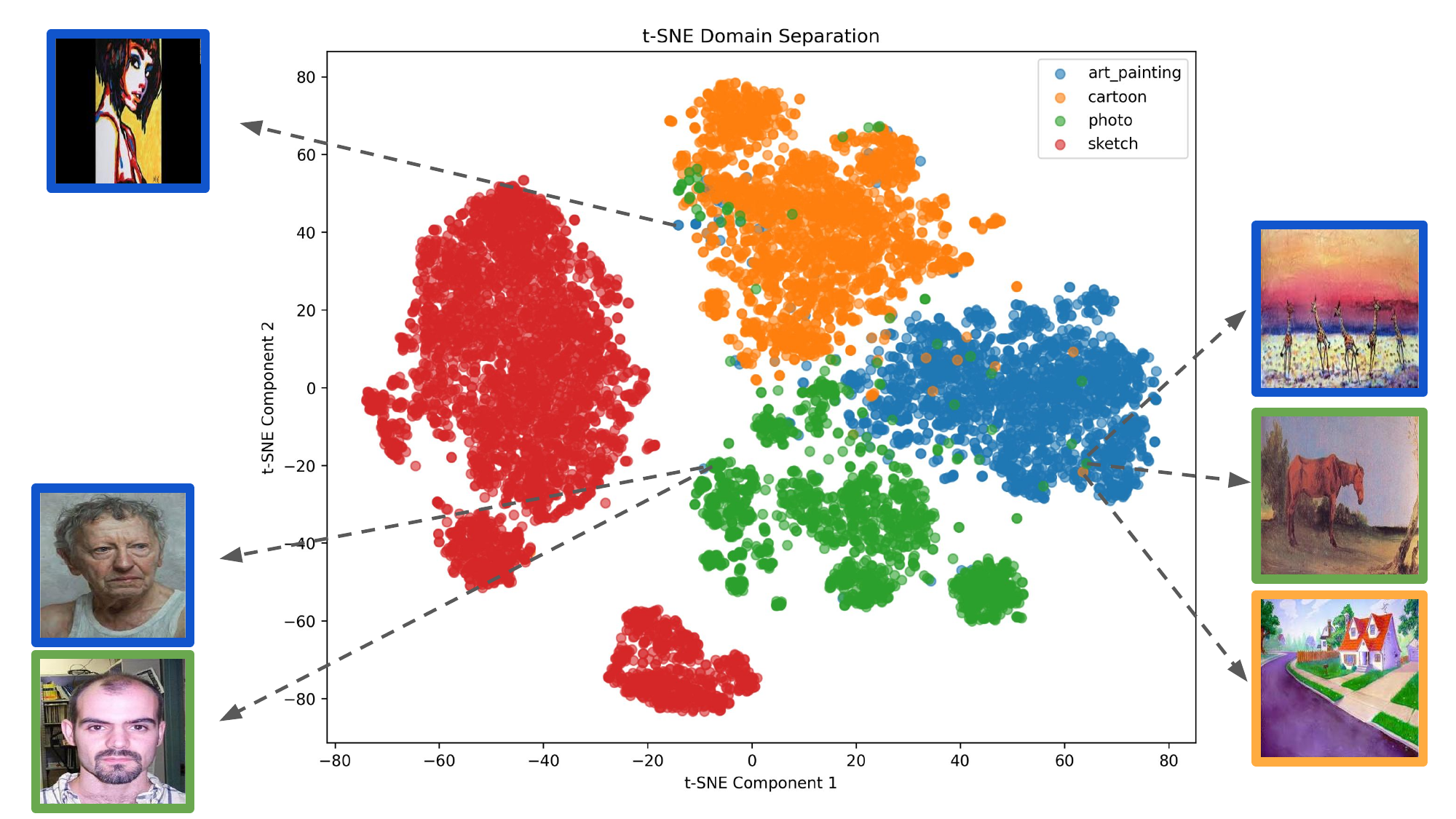}}\hfill \\ 
    \caption{\footnotesize{\textbf{Examples of inconsistent or confusing domain labels}. Given that most datasets in this study are web-scraped, we expect there to be label noise and domain inconsistencies which may impact the NMI scores. These examples from the PACS dataset and SD-2.1 feature space illustrate cases where domain assignments may be unclear or conflicting. The color of the border on the images denotes the ground truth domain label.}}
    \label{fig:adaptive_gap}
\end{figure} 
\FloatBarrier

\clearpage


\section{Effect of Text-Conditioning in SD-2.1 for Domain Separation}
\begin{table}[!h]
\centering
\resizebox{0.45\textwidth}{!}{%
\begin{tabular}{lcccc}
\toprule
\textbf{Dataset} & \multicolumn{2}{c}{\textbf{Domain NMI}} & \multicolumn{2}{c}{\textbf{Domain Predictability}} \\
\cmidrule(lr){2-3} \cmidrule(lr){4-5}
                 & \textbf{Empty Prompt} & \textbf{Prompt} & \textbf{Empty Prompt} & \textbf{Prompt} \\
\midrule
PACS            & 0.82   & 0.85   & 98.95   & 99.51   \\
OfficeHome      & 0.22   & 0.24   & 86.43   & 92.91   \\
\bottomrule
\end{tabular}}
\caption{\footnotesize \textbf{Domain NMI and predictability scores for empty vs text conditioned prompts for SD-2.1 on PACS and OfficeHome.} For text conditioning we used the prompt: ``A photo of an object in the style of \{domain\}". Similar to the findings of \citet{revelio}, text conditioning appears to activate more relevant features.}
\label{tab:prompt_comparison}
\end{table}

\section{Effect of Layer and Timestep in Diffusion Models for Domain Separation (DiT vs SD-2.1) on PACS, and VLCS}

\begin{figure}[!h]
    \centering
    \includegraphics[width=0.45\linewidth]{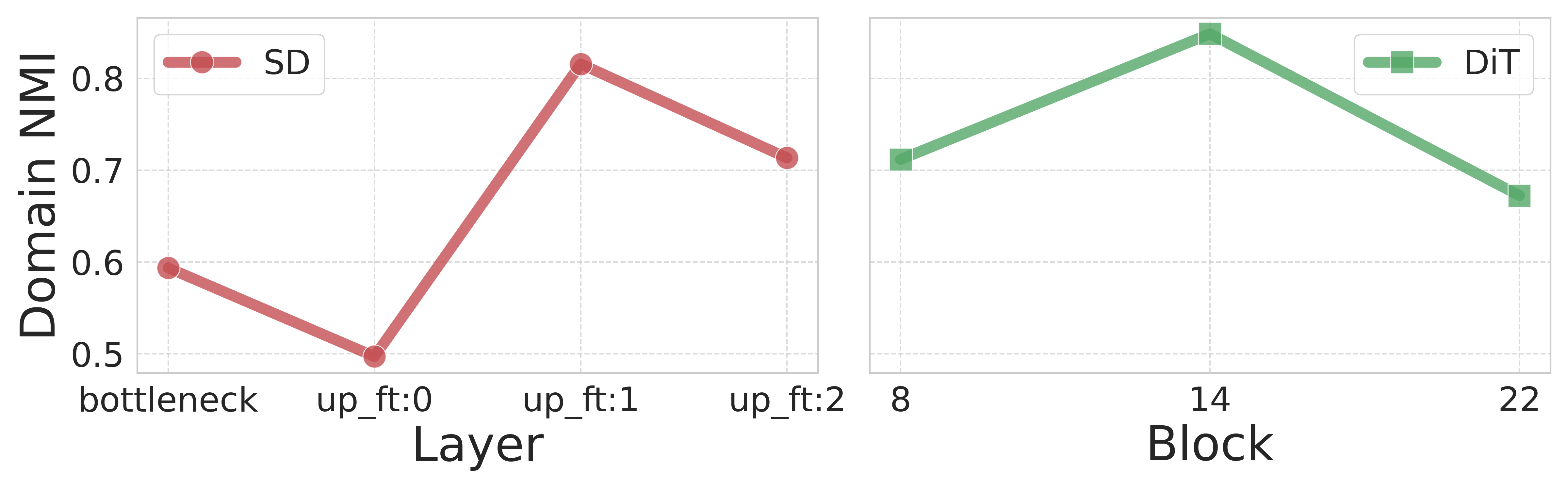} \\
    \vspace{0.1cm} 

    \includegraphics[width=0.65\linewidth]{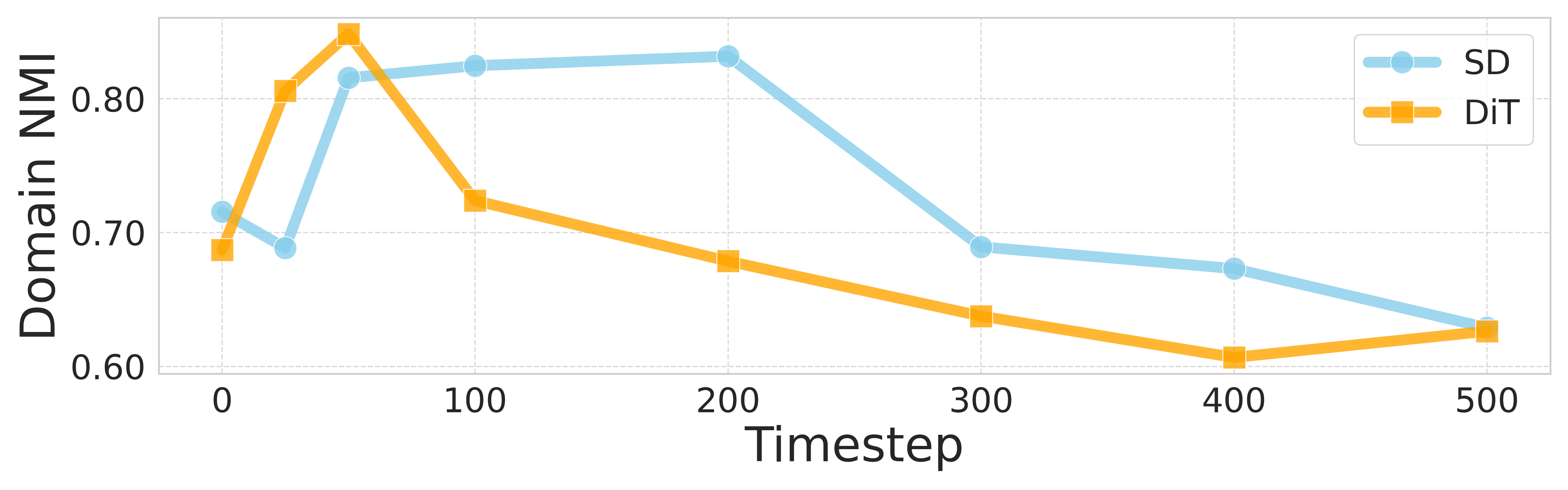}
    
    \caption{\footnotesize{\textbf{Domain NMI comparison across layers and timesteps for PACS.} 
    Top: Domain NMI scores for SD-2.1 layers (best: up\_ft:1) and DiT blocks (best: block:14). Bottom: Domain NMI scores across various denoising timesteps for SD-2.1 and DiT on PACS.}}
    \label{fig:domain_nmi_grid}
\end{figure}

\begin{figure}[!h]
    \centering
    \includegraphics[width=0.45\linewidth]{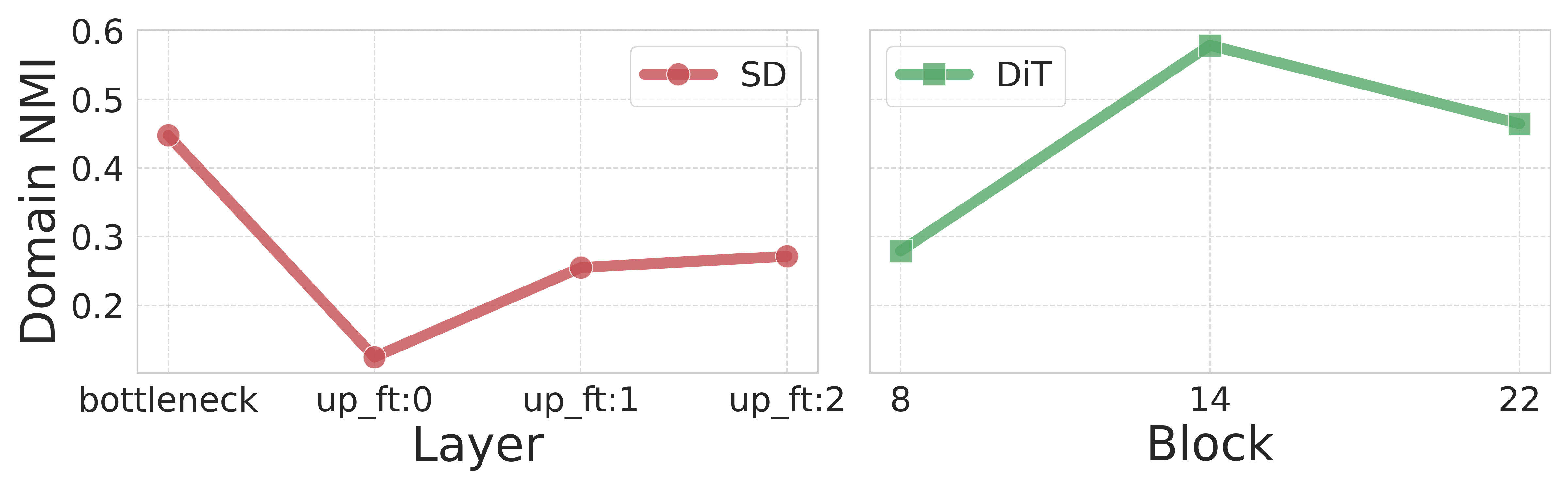} \\
    \caption{\footnotesize{\textbf{Domain NMI comparison across layers for VLCS.} The Bottleneck Layer of Stable Diffusion (SD-2.1) which capture more coarse-grained features aids in separating high-level domain shifts in VLCS. However, DiT's superior capability to capture global context via self-attention outperforms the domain NMI scores at \texttt{bottleneck} and \texttt{up\_ft:1}.}}
    \label{fig:vlcs_domain_nmi_grid}
\end{figure}

Following ~\citet{revelio}, we choose a lower noise level at timestep (t=$50$), with a motivation to capture rich fine-grained visual information. We use t=$50$ for both DiT (at block 14) and SD-2.1 (at \texttt{up\_ft:1}) for both class and domain NMI scores (in Tables~\ref{tab:dataset_comparison}, and~\ref{tab:class_nmi_dataset_comparison}), and to obtain the classification accuracies in Table~\ref{tab:domainbed}. In Fig.~\ref{fig:domain_nmi_grid}, we observe that t=$50$ provides the highest domain NMI score for PACS using DiT. We also note that on VLCS, the \texttt{bottleneck} layer outperforms the domain NMI score obtained from \texttt{up\_ft:1} in Fig.~\ref{fig:vlcs_domain_nmi_grid}, likely due it's focus on coarse-grained features as noted in~\cite{revelio}.

\FloatBarrier
\clearpage


\clearpage
\section{{\algoname} Pseudo-code}

\begin{algorithm}[!h]
\footnotesize
\caption{Training Pseudocode with RBF Kernel Ridge Regression} 
\textbf{Input:} Training data $D_{\text{tr}}$, transform schedule $T_{\text{transform}}$, $K$: \#clusters \\
\textbf{Output:} $F_{\text{image}}(.; \bomega)$, $F_{\text{MLP}}(.; \bW)$, mapping $\mathcal{T}$

\begin{algorithmic}
\STATE \textbf{Initialize:} Compute feature representations $\bPsi, \bPhi$, initialize model parameters $\bomega_0, \bW$.
\STATE $\{\bpsi_k\}, \{D_k\} \leftarrow \textsc{Clustering}(\bPsi, K)$ 
\FOR{$t = 1$ to $T$}
    \IF{$t \in T_{\text{transform}}$} 
        \STATE \textbf{For each} $k$: $\widehat{\bPhi}_k = \frac{1}{|D_k|}\sum_{\x \in D_k}\bPhi(\x)$ 
        \STATE Compute pairwise distances $\|\bpsi_i - \bpsi_j\|_2, \forall i \neq j$
        \STATE $\gamma \leftarrow 1 / (2 \cdot \text{median}(\text{pairwise distances})^2)$ \COMMENT{using median heuristic}
        \STATE \textbf{Fit} $\mathcal{T}$ \textbf{via RBF Kernel Ridge Regression} using $\{\widehat{\bpsi}_k\} \mapsto \{\widehat{\bPhi}_k\}$ and $\gamma$
        \STATE $\bpsi_\x' \leftarrow \mathcal{T}(\bpsi_\x)$
    \ENDIF
    
    \FOR{batch $(\x, \bpsi_\x, y)$ in $D_{\text{tr}}$}
        \STATE $\bPhi(\x) \leftarrow F_{\text{image}}(\x; \bomega_{t})$ 
        \STATE $\bpsi_\x' \leftarrow \mathcal{T}(\bpsi_\x)$
        \STATE $\hat{y} \leftarrow F_{\text{MLP}}\bigl(\textsc{Concat}\bigl(\bPhi(\x), \bpsi_\x'\bigr); \bW_t\bigl)$
        \STATE Update $\bomega_{t+1}, \bW_{t+1}$ via \textsc{SGD Step} on $\mathcal{L} = \textsc{CrossEntropy}(\hat{y}, y)$
    \ENDFOR
\ENDFOR
\STATE \textbf{Return} $F_{\text{image}}(.; \bomega_T)$, $F_{\text{MLP}}(.; \bW_T)$, and $\mathcal{T}$
\end{algorithmic}

\noindent\hrulefill

\textbf{Inference}\\
\textbf{Input:} Test data $D_{\text{test}}$, transformation function $\mathcal{T}$, and centroids $\{\widehat{\bpsi}_k\}_{k=1}^K$\\
\textbf{Output:} Predicted labels $\hat{y}$

\begin{algorithmic}
\FOR{$\x \in D_{\text{test}}$}
    \STATE $\bpsi_\x \leftarrow \textsc{NearestCentroid}\!\bigl(\bPsi, \x\bigr)$ 
        \COMMENT{Find closest cluster in \(\bPsi\)-space}
    \STATE $\bpsi_\x' \leftarrow \mathcal{T}(\bpsi_\x)$ 
        \COMMENT{Apply same RBF transform as in training}
    \STATE $\bPhi(\x) \leftarrow F_{\text{image}}(\x; \bomega_T)$
    \STATE $\hat{y} \leftarrow F_{\text{MLP}}\bigl(\textsc{Concat}(\bPhi(\x), \bpsi_\x'); \bW_T\bigl)$
\ENDFOR
\STATE \textbf{Return} $\hat{y}$
\end{algorithmic}
\label{alg:adaclustv2}
\end{algorithm}
\FloatBarrier
\clearpage

\section{Domain Shift Examples and Domain Separation in Feature Spaces}

In this section, we provide:
\begin{itemize}
    \item \textbf{Example images}, i.e. class samples across domains for each dataset.
    \item \textbf{Class vs Domain NMI scores} for each feature extractor ($\bPsi$) studied in this work, on each dataset.
    \item \textbf{Feature space visualizations} for each feature extractor ($\bPsi$) studied in this work, on the PACS, VLCS, OfficeHome, and TerraInognita datasets.
\end{itemize}


\subsection{PACS~\cite{PACS}}
\begin{figure}[h]
    \centering
    \includegraphics[width=0.75\textwidth]{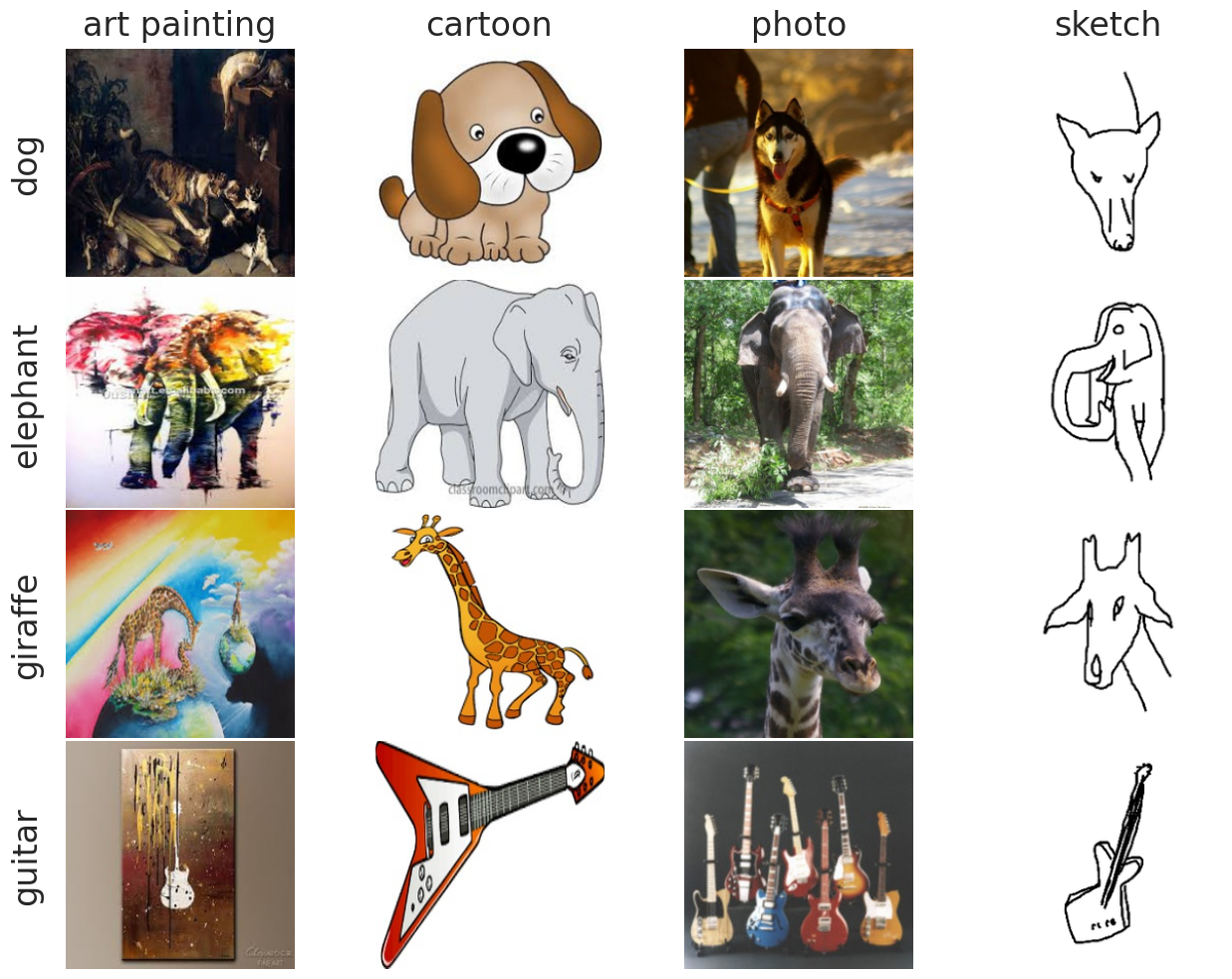}
    \caption{\footnotesize Class examples across domains in the PACS dataset. Each column represents a domain, and each row corresponds to a class.}
    \label{fig:PACS_grid}
\end{figure}
\FloatBarrier

\begin{table}[h]
    \centering
    \label{tab:PACS_classes_domains}
    \begin{tabular}{p{8cm}|p{8cm}}
        \hline
        \textbf{Domains} & \textbf{Classes} \\ 
        \hline
        art painting, cartoon, photo, sketch & dog, elephant, giraffe, guitar, horse, house, person \\ 
        \hline
    \end{tabular}
    \caption{\footnotesize 4 domains and 7 classes of the PACS dataset.}
\end{table}
\FloatBarrier

\begin{figure}[h]
    \centering
    \includegraphics[width=0.55\textwidth]{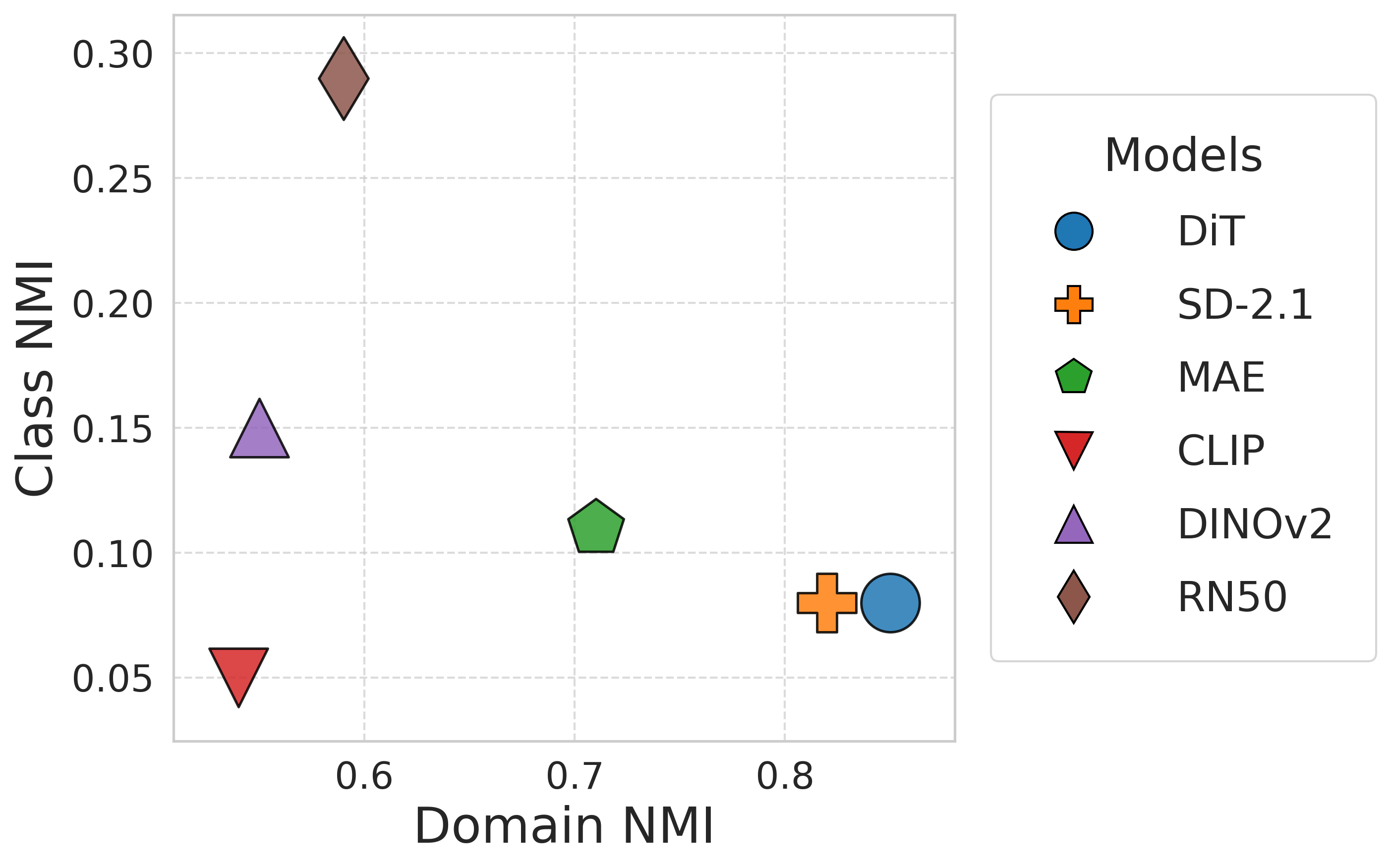}
    \caption{\textbf{Class vs Domain NMI scores for PACS.} Note how RN50 has the highest class NMI and diffusion models have low class NMI scores. Diffusion models also has the highest domain NMI scores, thereby capturing domain-specific class invariant structures.}
    \label{fig:OH_net_nmi_grid}
\end{figure}


\begin{figure}[h]
    \centering
    \includegraphics[width=0.75\textwidth]{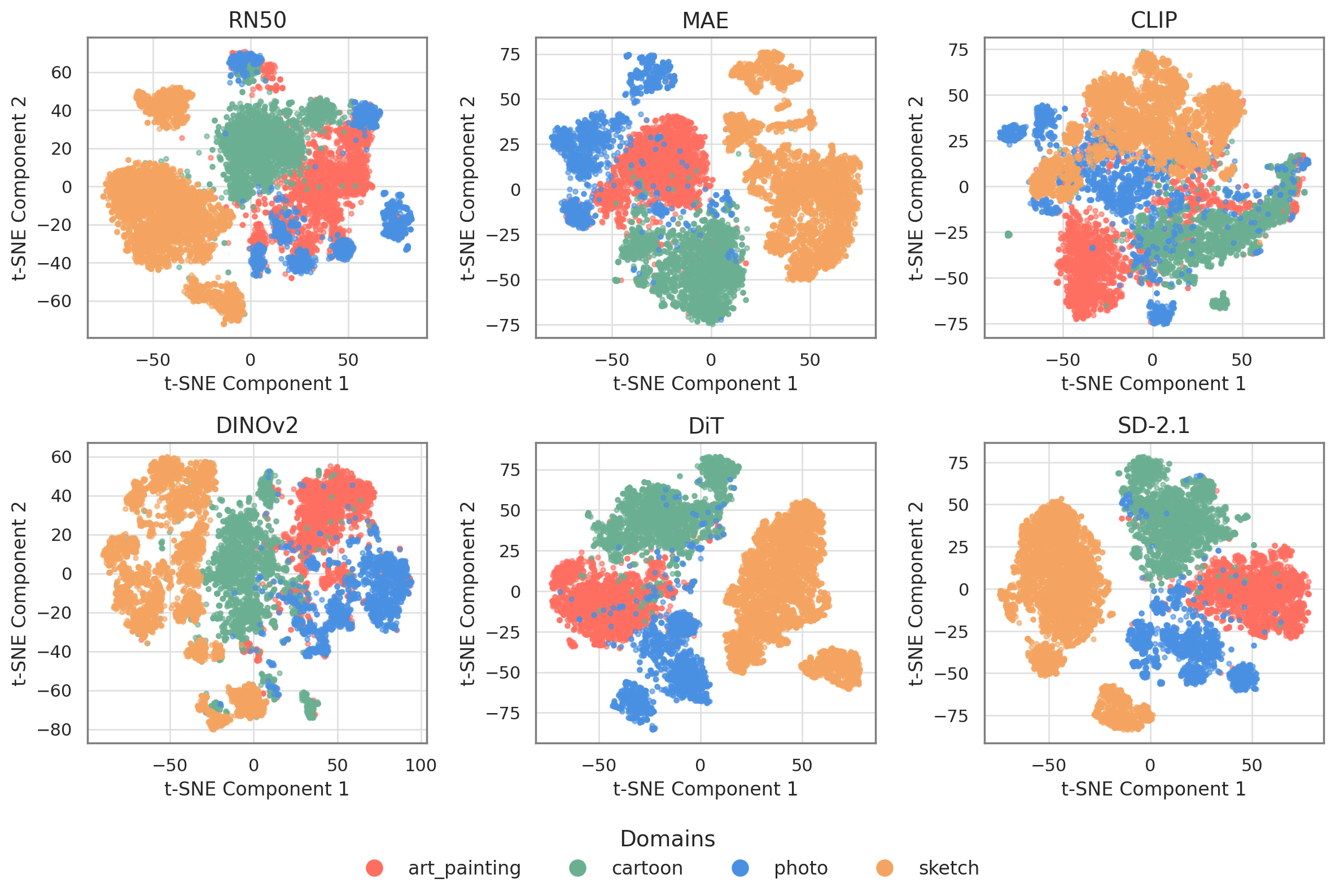}
    \caption{\footnotesize \textbf{T-SNE visualization of domain separation for PACS}. Each point represents a sample, colored by its domain. Notice how well separated the domains are when diffusion features are used compared to other models.}
    \label{fig:PACS_tsne}
\end{figure}


\clearpage
\subsection{VLCS~\cite{VLCS}}
\begin{figure}[h]
    \centering
    \includegraphics[width=0.75\textwidth]{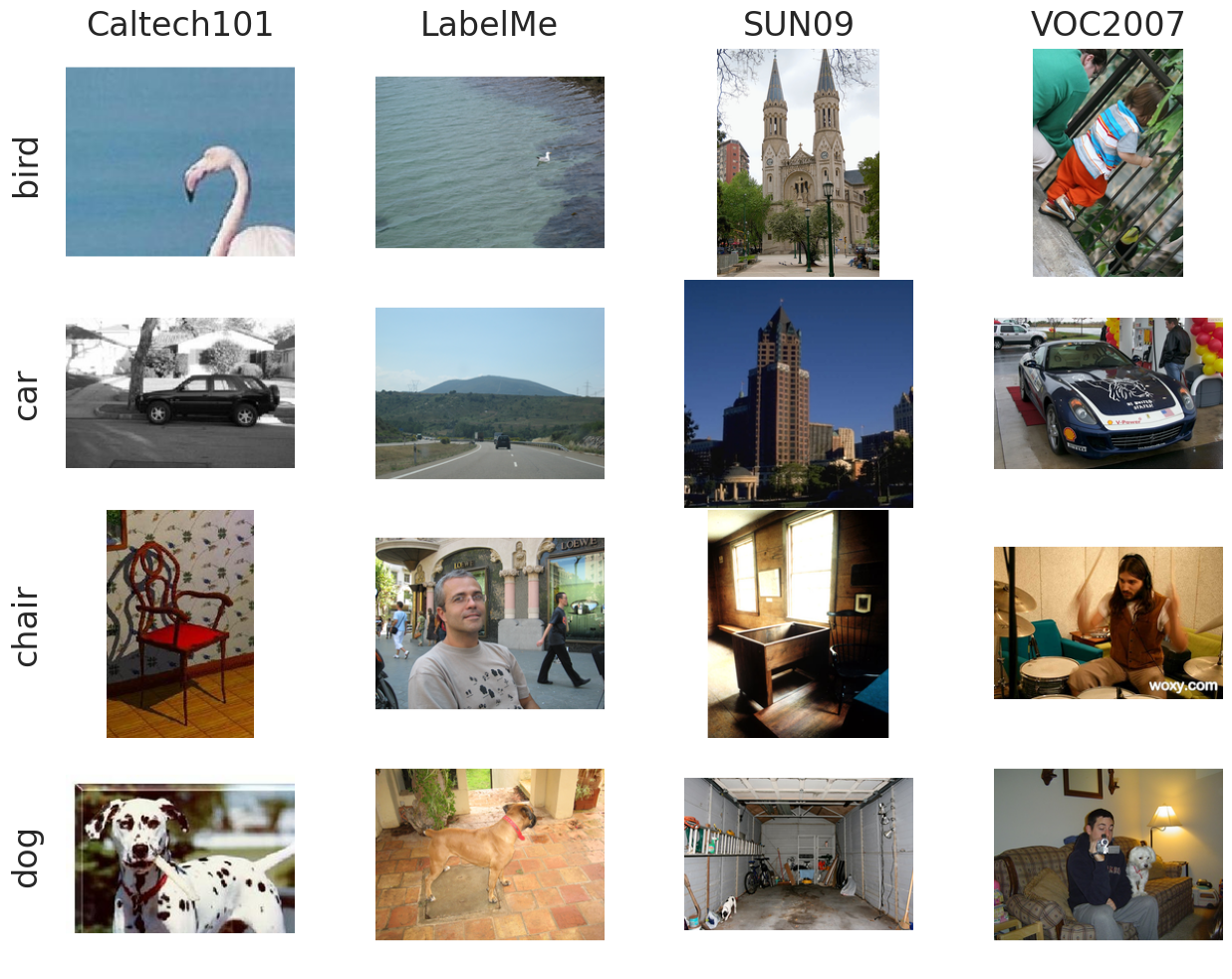}
    \caption{Class examples across domains in the VLCS dataset. Each column represents a domain, and each row corresponds to a class.}
    \label{fig:VLCS_grid}
\end{figure}
\FloatBarrier

\begin{table}[h]
    \centering
    \label{tab:VLCS_classes_domains}
    \begin{tabular}{p{8cm}|p{8cm}}
        \hline
        \textbf{Domains} & \textbf{Classes} \\ 
        \hline
        Caltech101, LabelMe, SUN09, VOC2007 & bird, car, chair, dog, person \\ 
        \hline
    \end{tabular}
    \caption{\footnotesize 4 domains and 5 classes of the VLCS dataset.}
\end{table}

\begin{figure}[h]
    \centering
    \includegraphics[width=0.55\textwidth]{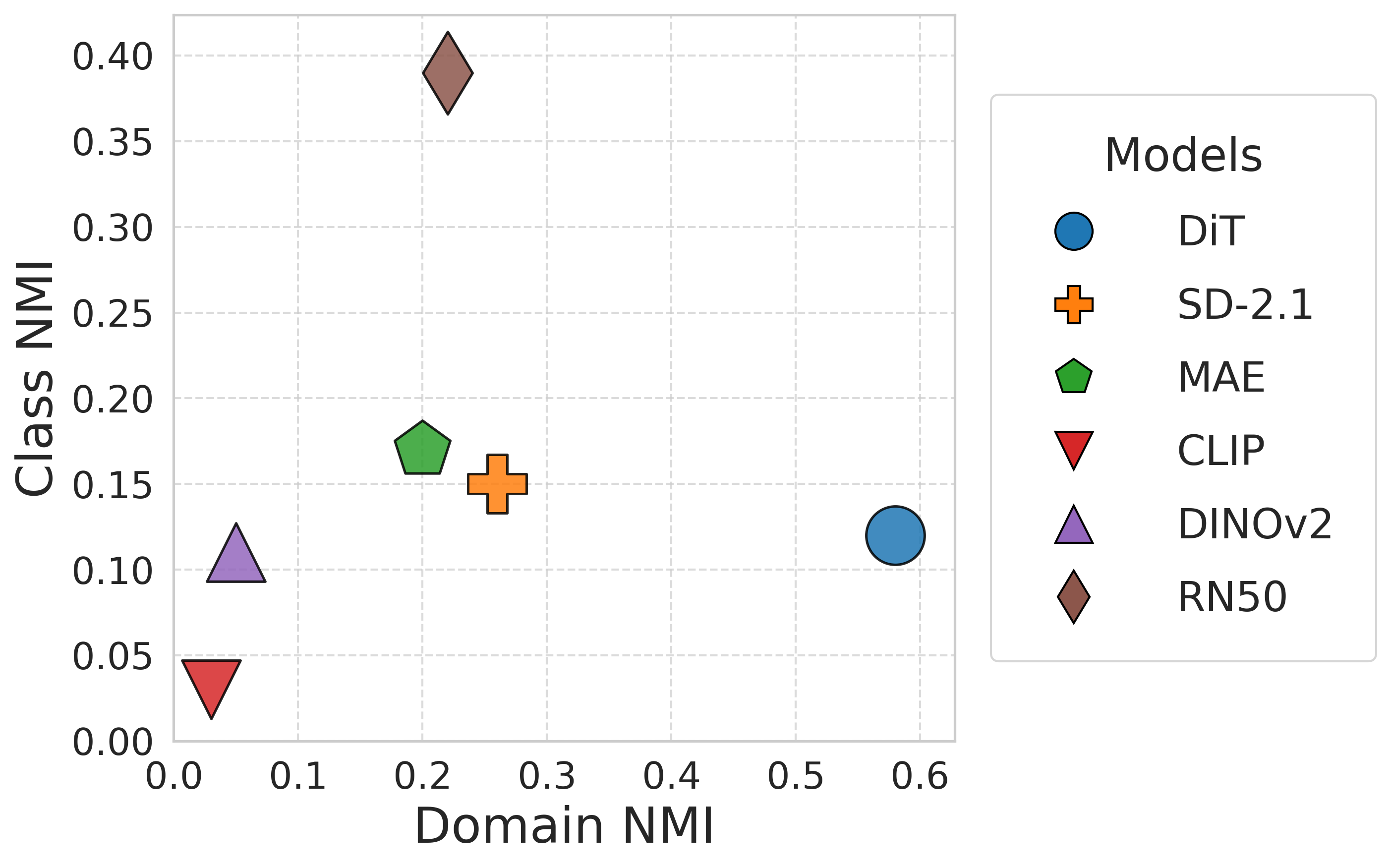}
    \caption{\textbf{Class vs Domain NMI scores for VLCS.} Note how RN50 has the highest class NMI score, and diffusion models have low class NMI scores. DiT has a much higher domain NMI score than SD-2.1, resulting from its stronger capability in capturing high-level dataset-specific biases, as discussed in Sec.~\ref{sec:pretrain_obj}. }
    \label{fig:domain_net_nmi_grid}
\end{figure}


\begin{figure}[h]
    \centering
    \includegraphics[width=0.75\textwidth]{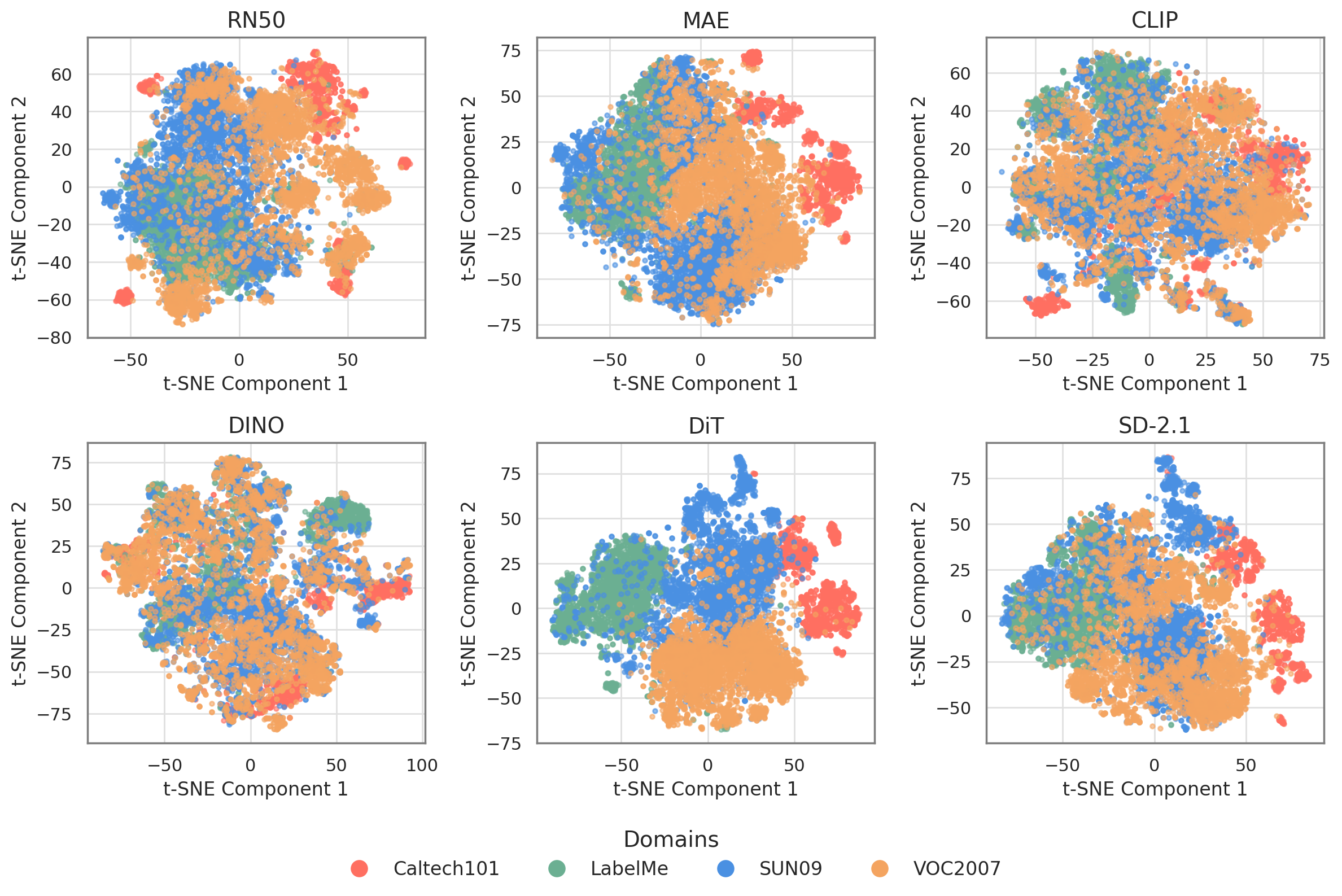}
    \caption{\textbf{T-SNE visualization of domain separation for VLCS}. Each point represents a sample, colored by its domain. Note how the DiT feature space best separate the domains.}
    \label{fig:PACS_tsne}
\end{figure}


\clearpage
\subsection{OfficeHome~\cite{OfficeHome}}
\begin{figure}[h]
    \centering
    \includegraphics[width=0.75\textwidth]{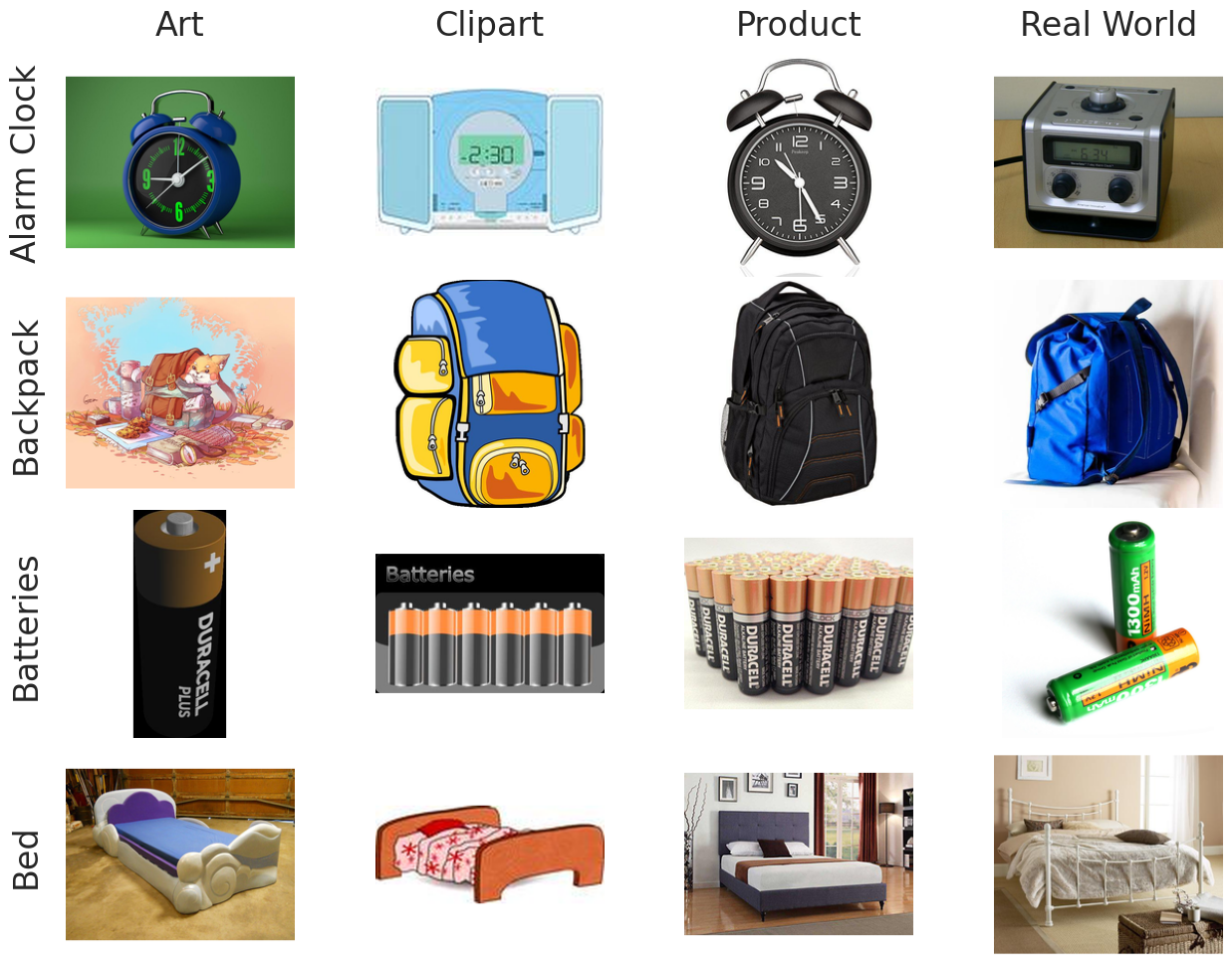}
    \caption{Class examples across domains in the OfficeHome dataset. Each column represents a domain, and each row corresponds to a class.}
    \label{fig:OfficeHome_grid}
\end{figure}

\begin{table}[h]
    \centering
    \renewcommand{\arraystretch}{1.1} 
    \setlength{\tabcolsep}{5pt} 

    \begin{tabular}{p{8cm}|p{8cm}} 
        \hline
        \textbf{Domains} & \textbf{Classes} \\ 
        \hline
        Art, Clipart, Product, Real World & 
        Alarm Clock, Backpack, Batteries, Bed, Bike, Bottle, Bucket, Calculator, Calendar, Candles, Chair, Clipboards, Computer, Couch, Curtains, Desk Lamp, Drill, Eraser, Exit Sign, Fan, File Cabinet, Flipflops, Flowers, Folder, Fork, Glasses, Hammer, Helmet, Kettle, Keyboard, Knives, Lamp Shade, Laptop, Marker, Monitor, Mop, Mouse, Mug, Notebook, Oven, Pan, Paper Clip, Pen, Pencil, Post-it Notes, Printer, Push Pin, Radio, Refrigerator, Ruler, Scissors, Screwdriver, Shelf, Sink, Sneakers, Soda, Speaker, Spoon, TV, Table, Telephone, ToothBrush, Toys, Trash Can, Webcam. \\ 
        \hline
    \end{tabular}

    \caption{\footnotesize 4 domains and 65 Classes of the OfficeHome dataset.}
    \label{tab:OH_classes_domains}
\end{table}

\begin{figure}[h]
    \centering
    \includegraphics[width=0.55\textwidth]{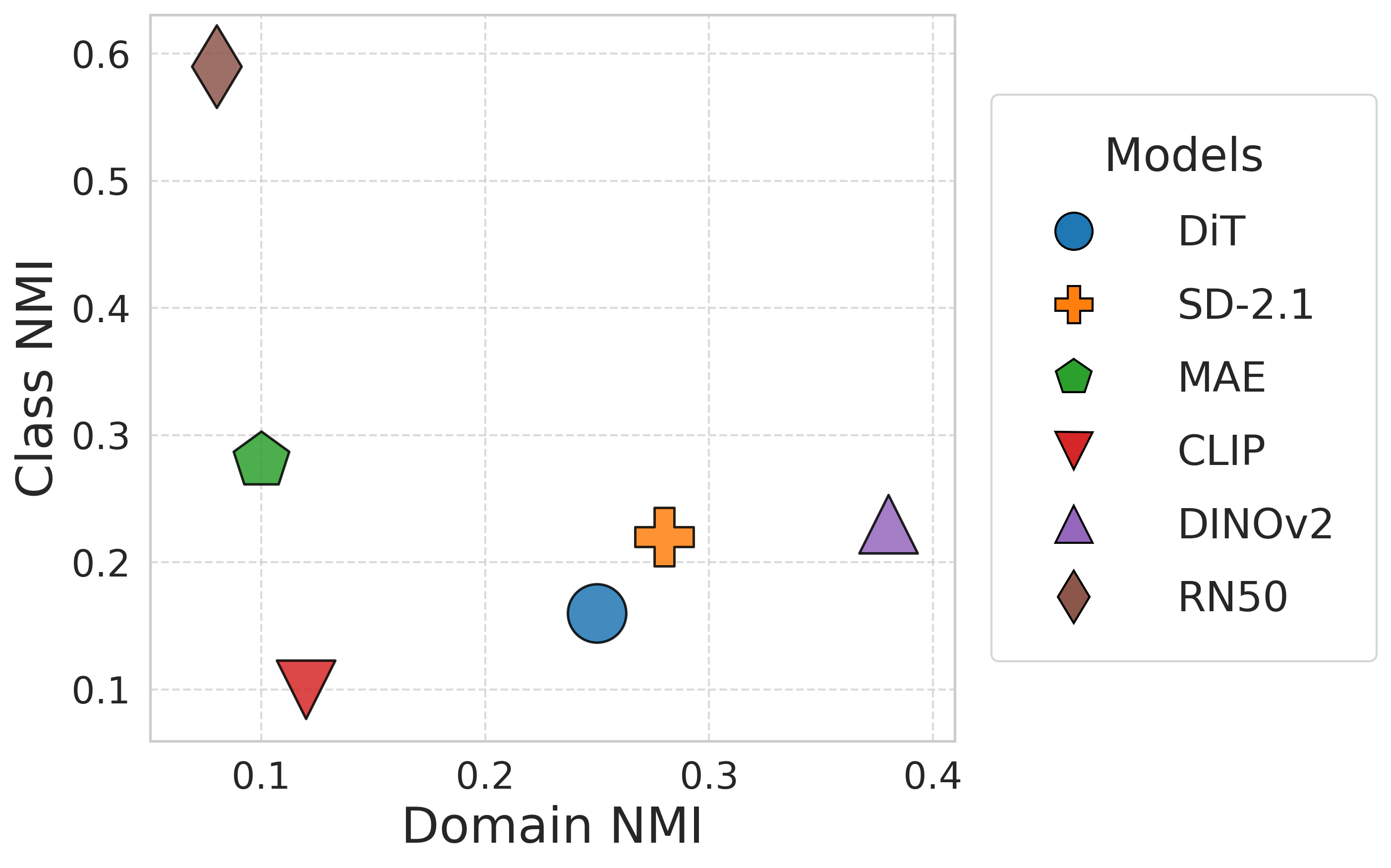}
    \caption{\textbf{Class vs Domain NMI scores for OfficeHome.} Note how RN50 has the highest class NMI score and DINOv2 has the highest domain NMI score, resulting form its stronger ability in capturing low-level style shifts, as discussed in Sec.~\ref{sec:pretrain_obj}. DiT and SD-2.1 have moderate domain NMI scores, with DiT having a lower class NMI score.}
    \label{fig:OH_net_nmi_grid}
\end{figure}


\begin{figure}[h]
    \centering
    \includegraphics[width=0.75\textwidth]{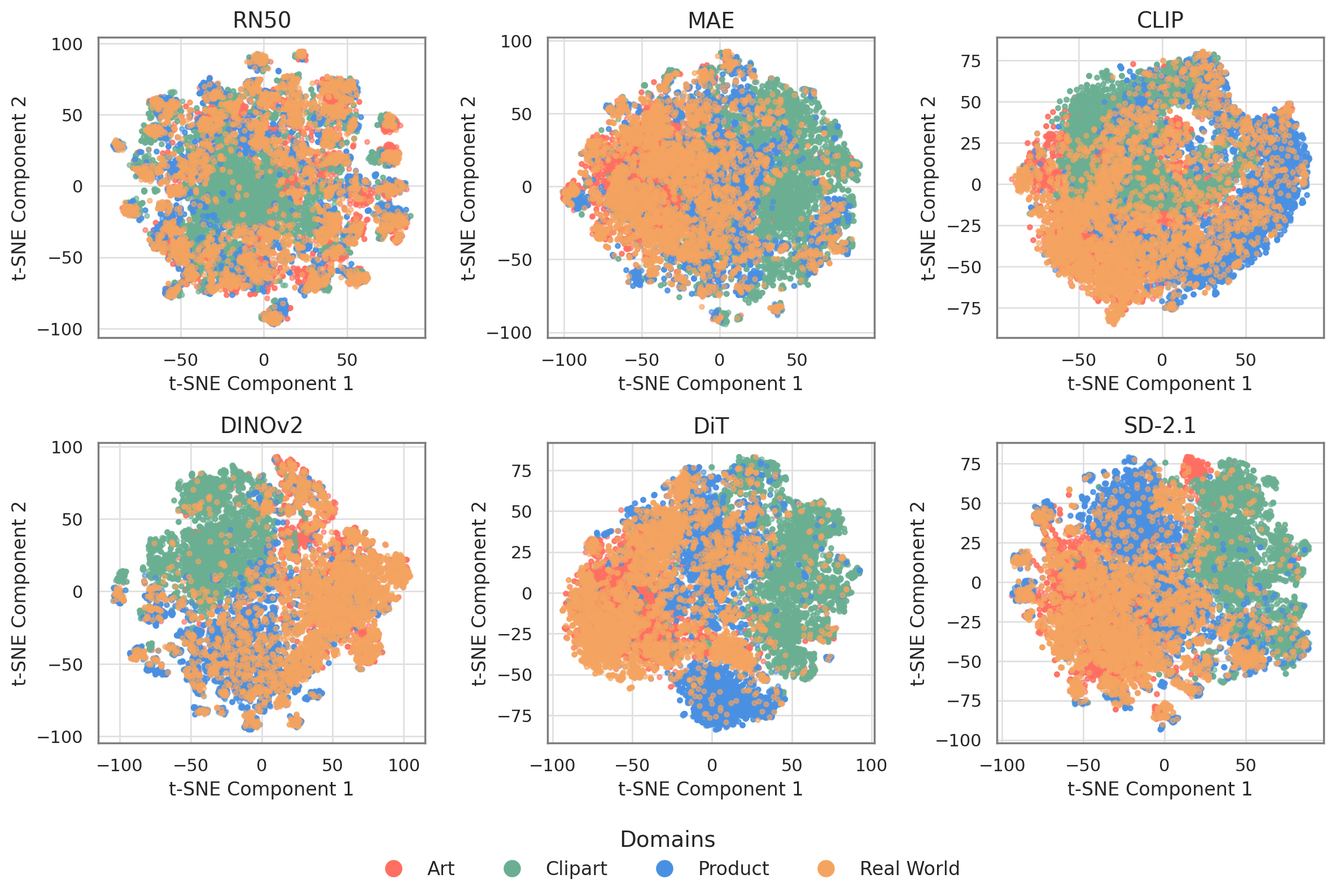}
    \caption{\textbf{T-SNE visualization of domain separation for OfficeHome.} Each point represents a sample, colored by its domain. All models struggle to separate the domains in this dataset. The ``real" domain has considerable overlap with the other domains.}
    \label{fig:OH_tsne}
\end{figure}




\clearpage
\subsection{TerraIncognita~\cite{TerraInc}}
\begin{figure}[h]
    \centering
    \includegraphics[width=0.75\textwidth]{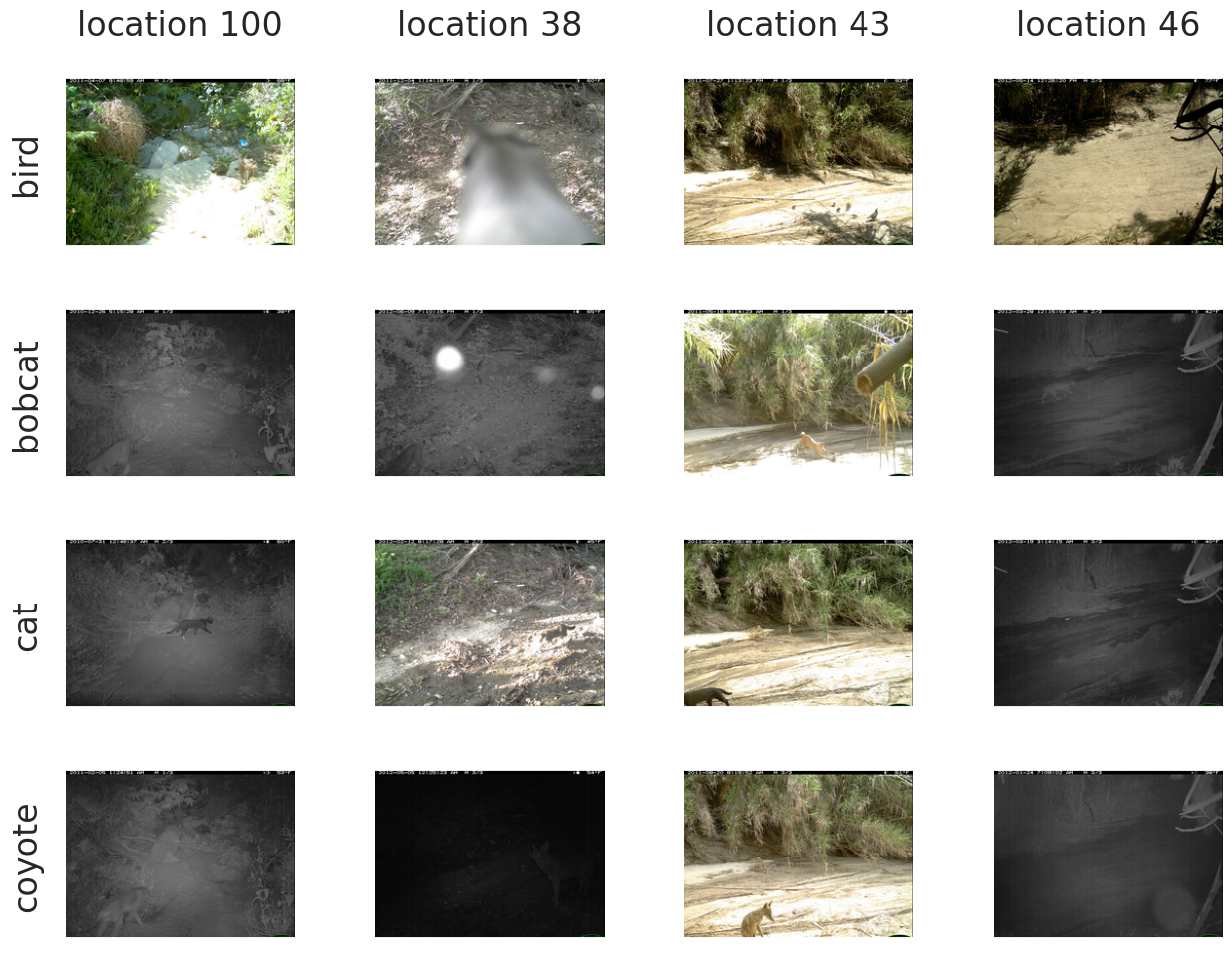}
    \caption{Class examples across domains in the TerraIncognita dataset. Each column represents a domain, and each row corresponds to a class.}
    \label{fig:TerraIncognita_grid}
\end{figure}

\begin{table}[h]
    \centering
    \renewcommand{\arraystretch}{1.1} 
    \setlength{\tabcolsep}{5pt} 

    \begin{tabular}{p{8cm}|p{8cm}} 
        \hline
        \textbf{Domains} & \textbf{Classes} \\ 
        \hline
        Location 100, Location 38, Location 43, Location 46 & 
        bird, bobcat, cat, coyote, dog, empty, opossum, rabbit, raccoon, squirrel \\ 
        \hline
    \end{tabular}

    \caption{\footnotesize 4 domains and 10 classes of the TerraIncognita dataset.}
    \label{tab:Terra_classes_domains}
\end{table}

\begin{figure}[h]
    \centering
    \includegraphics[width=0.55\textwidth]{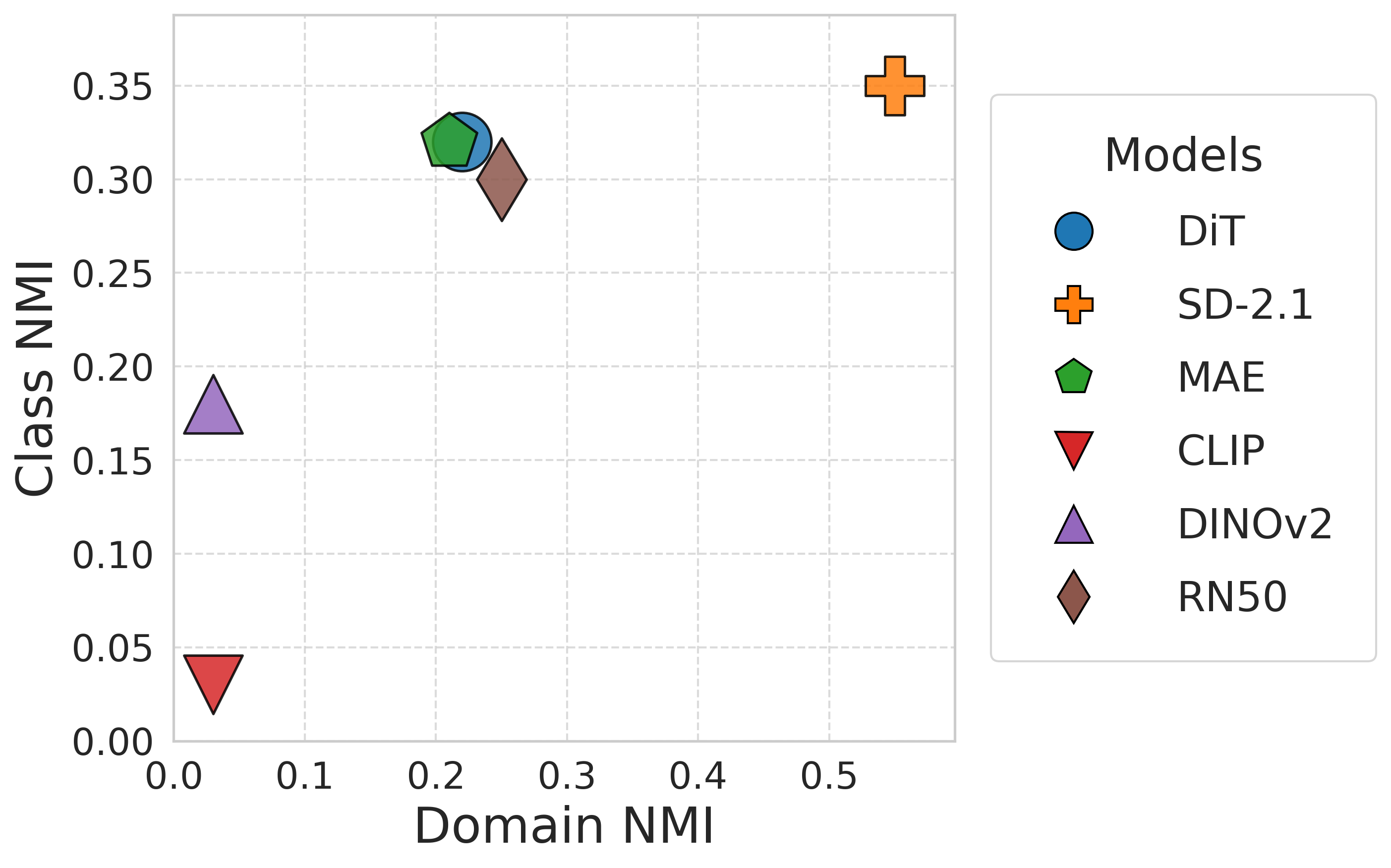}
    \caption{\textbf{Class vs Domain NMI scores for TerraIncognita.} Most models have a high class NMI score. SD-2.1 has the highest domain NMI score, resulting from its stronger capability in capturing spatial information, as discussed in Sec.~\ref{sec:pretrain_obj}.}
    \label{fig:domain_net_nmi_grid}
\end{figure}


\begin{figure}[h]
    \centering
    \includegraphics[width=0.75\textwidth]{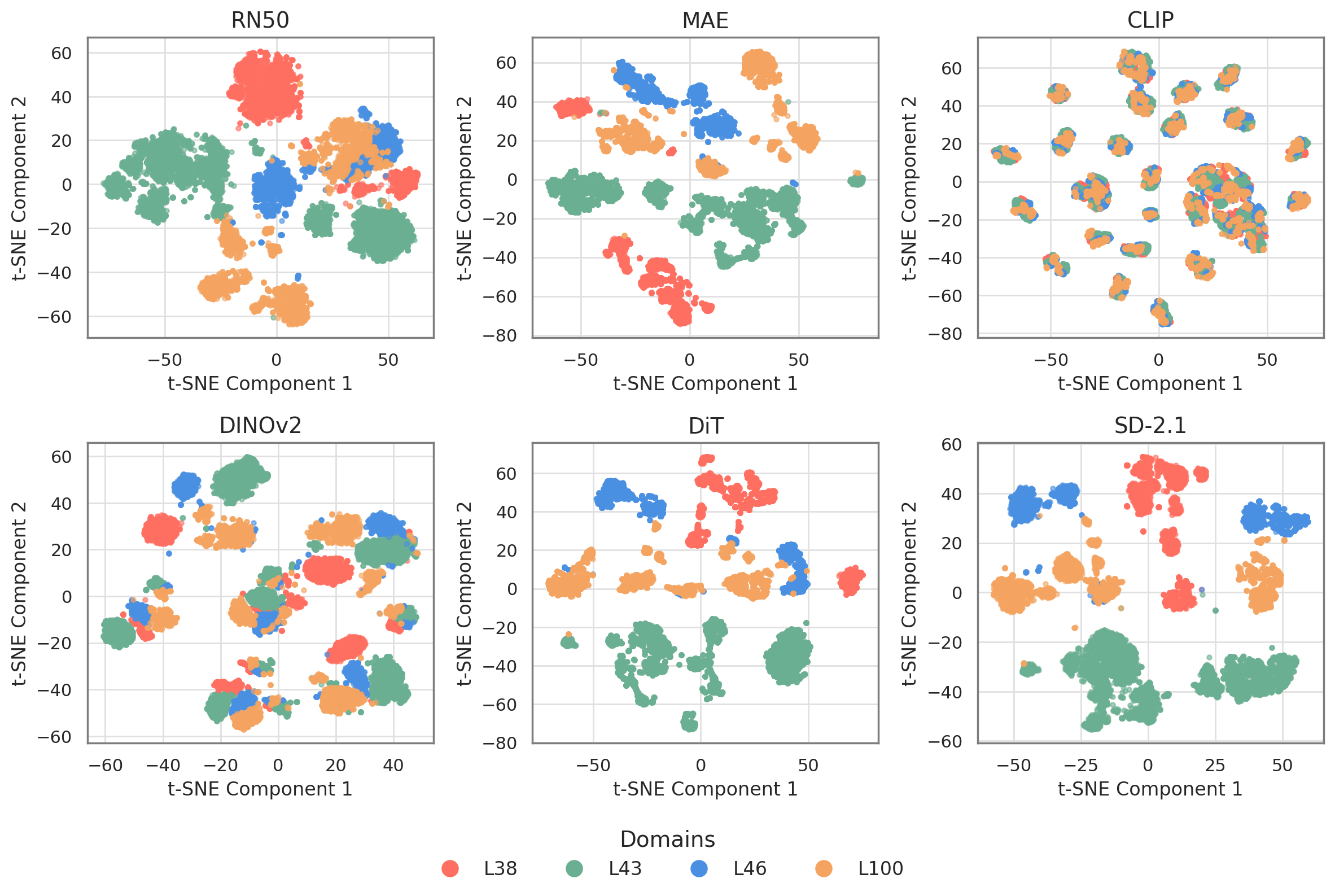}
    \caption{\textbf{T-SNE visualization of domain separation for TerraIncognita}. Each point represents a sample, colored by its domain. Note how the SD-2.1 feature space best groups samples from the same domain closer together, and separate from other domains.}
    \label{fig:Terra_tsne}
\end{figure}



\clearpage
\subsection{DomainNet~\cite{domainnet}}
\begin{figure}[h]
    \centering
    \includegraphics[width=0.85\textwidth]{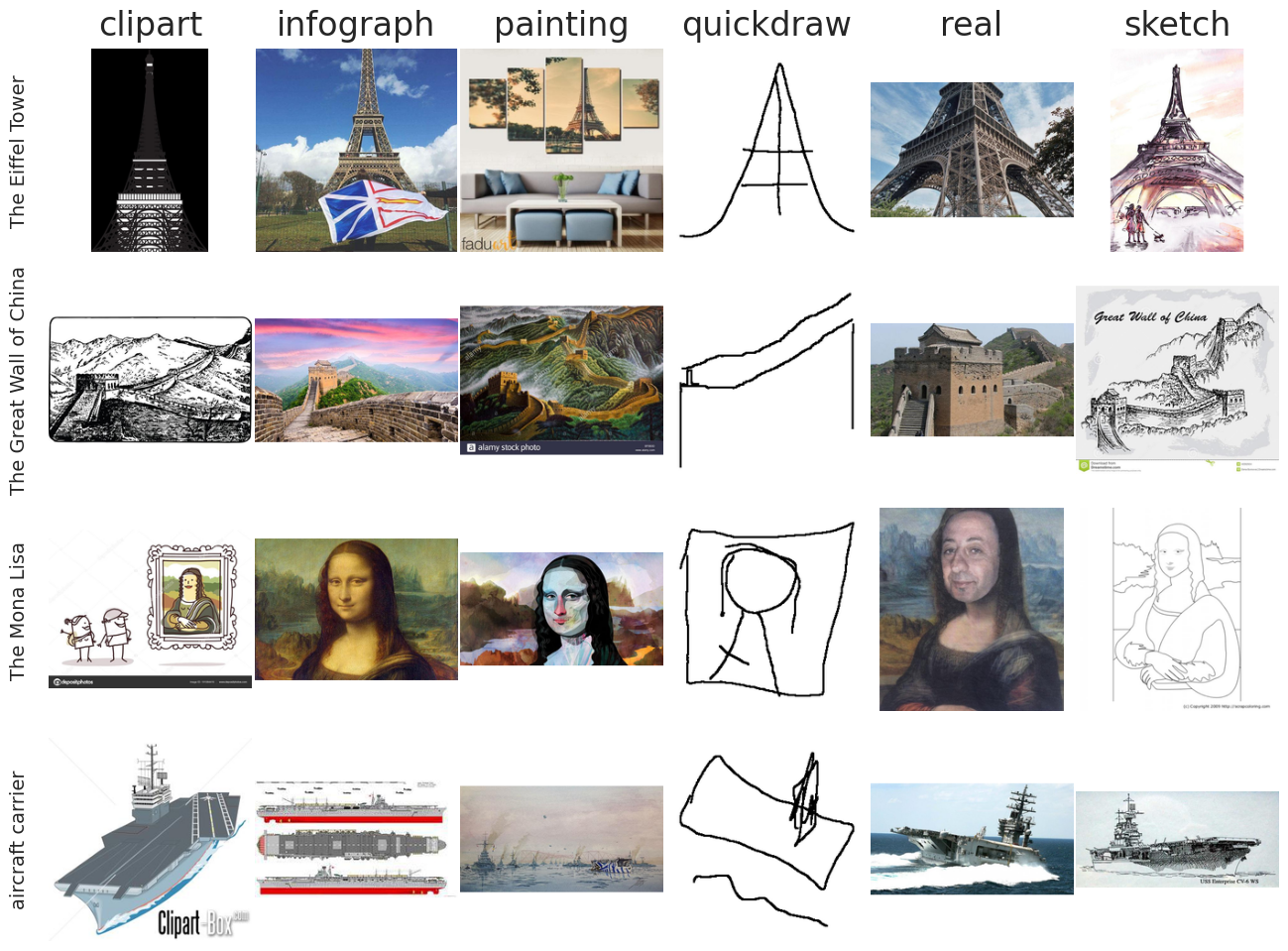}
    \caption{Class examples across domains in the DomainNet dataset. Each column represents a domain, and each row corresponds to a class.}
    \label{fig:domain_net_grid}
\end{figure}

\begin{figure}[h]
    \centering
    \includegraphics[width=0.55\textwidth]{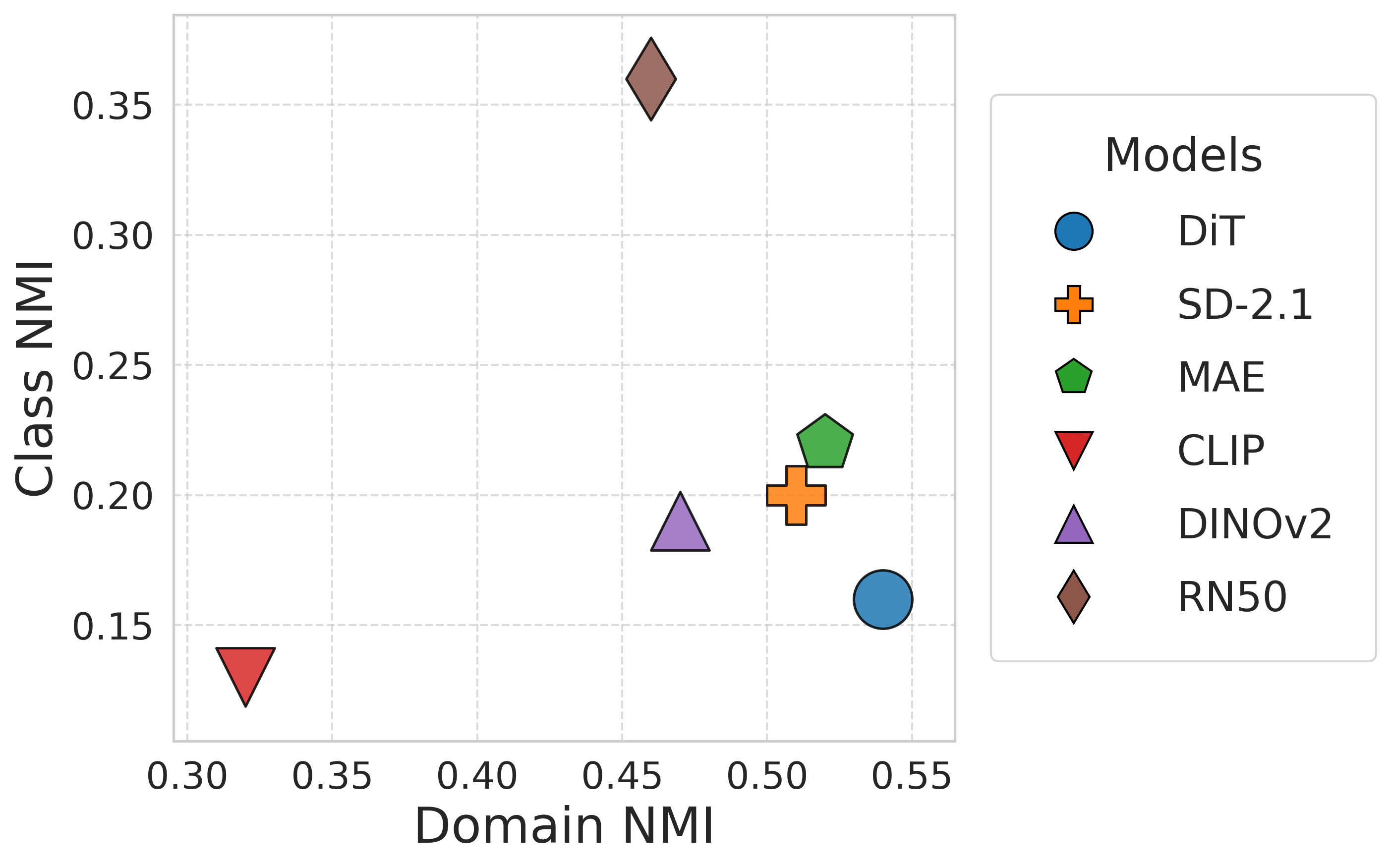}
    \caption{\textbf{Class vs Domain NMI scores for DomainNet.} Note how RN50 has the highest class NMI and diffusion models, and MAE have the highest domain NMI scores, with DiT having a lower class NMI score. All models except CLIP exhibit a moderate domain NMI score, likely due to the varied domain shifts inherent in the dataset, as discussed in Sec.~\ref{sec:pretrain_obj}.}
    \label{fig:domain_net_nmi_grid}
\end{figure}

\clearpage
\begin{table}[h]
    \centering
    \renewcommand{\arraystretch}{1.1} 
    \setlength{\tabcolsep}{5pt} 

    \begin{tabular}{p{8cm}|p{9cm}} 
        \hline
        \textbf{Domains} & \textbf{Classes} \\ 
        \hline
        clipart, infograph, painting, quickdraw, real, sketch & 
        The Eiffel Tower, The Great Wall of China, The Mona Lisa, aircraft carrier, airplane, alarm clock, ambulance, angel, animal migration, ant, anvil, apple, arm, asparagus, axe, backpack, banana, bandage, barn, baseball, baseball bat, basket, basketball, bat, bathtub, beach, bear, beard, bed, bee, belt, bench, bicycle, binoculars, bird, birthday cake, blackberry, blueberry, book, boomerang, bottlecap, bowtie, bracelet, brain, bread, bridge, broccoli, broom, bucket, bulldozer, bus, bush, butterfly, cactus, cake, calculator, calendar, camel, camera, camouflage, campfire, candle, cannon, canoe, car, carrot, castle, cat, ceiling fan, cell phone, cello, chair, chandelier, church, circle, clarinet, clock, cloud, coffee cup, compass, computer, cookie, cooler, couch, cow, crab, crayon, crocodile, crown, cruise ship, cup, diamond, dishwasher, diving board, dog, dolphin, donut, door, dragon, dresser, drill, drums, duck, dumbbell, ear, elbow, elephant, envelope, eraser, eye, eyeglasses, face, fan, feather, fence, finger, fire hydrant, fireplace, firetruck, fish, flamingo, flashlight, flip flops, floor lamp, flower, flying saucer, foot, fork, frog, frying pan, garden, garden hose, giraffe, goatee, golf club, grapes, grass, guitar, hamburger, hammer, hand, harp, hat, headphones, hedgehog, helicopter, helmet, hexagon, hockey puck, hockey stick, horse, hospital, hot air balloon, hot dog, hot tub, hourglass, house, house plant, hurricane, ice cream, jacket, jail, kangaroo, key, keyboard, knee, knife, ladder, lantern, laptop, leaf, leg, light bulb, lighter, lighthouse, lightning, line, lion, lipstick, lobster, lollipop, mailbox, map, marker, matches, megaphone, mermaid, microphone, microwave, monkey, moon, mosquito, motorbike, mountain, mouse, moustache, mouth, mug, mushroom, nail, necklace, nose, ocean, octagon, octopus, onion, oven, owl, paint can, paintbrush, palm tree, panda, pants, paper clip, parachute, parrot, passport, peanut, pear, peas, pencil, penguin, piano, pickup truck, picture frame, pig, pillow, pineapple, pizza, pliers, police car, pond, pool, popsicle, postcard, potato, power outlet, purse, rabbit, raccoon, radio, rain, rainbow, rake, remote control, rhinoceros, rifle, river, roller coaster, rollerskates, sailboat, sandwich, saw, saxophone, school bus, scissors, scorpion, screwdriver, sea turtle, see saw, shark, sheep, shoe, shorts, shovel, sink, skateboard, skull, skyscraper, sleeping bag, smiley face, snail, snake, snorkel, snowflake, snowman, soccer ball, sock, speedboat, spider, spoon, spreadsheet, square, squiggle, squirrel, stairs, star, steak, stereo, stethoscope, stitches, stop sign, stove, strawberry, streetlight, string bean, submarine, suitcase, sun, swan, sweater, swing set, sword, syringe, t-shirt, table, teapot, teddy-bear, telephone, television, tennis racquet, tent, tiger, toaster, toe, toilet, tooth, toothbrush, toothpaste, tornado, tractor, traffic light, train, tree, triangle, trombone, truck, trumpet, umbrella, underwear, van, vase, violin, washing machine, watermelon, waterslide, whale, wheel, windmill, wine bottle, wine glass, wristwatch, yoga, zebra, zigzag \\ 
        \hline
    \end{tabular}

    \caption{\footnotesize 6 domains and 325 classes of the DomainNet dataset.}
    \label{tab:DN_classes_domains}
\end{table}
\clearpage


\FloatBarrier

\clearpage
\section{Synth-Photography and Synth-Artists Custom Datasets}

\begin{figure}[H]
    \centering
    \includegraphics[width=0.60\textwidth]{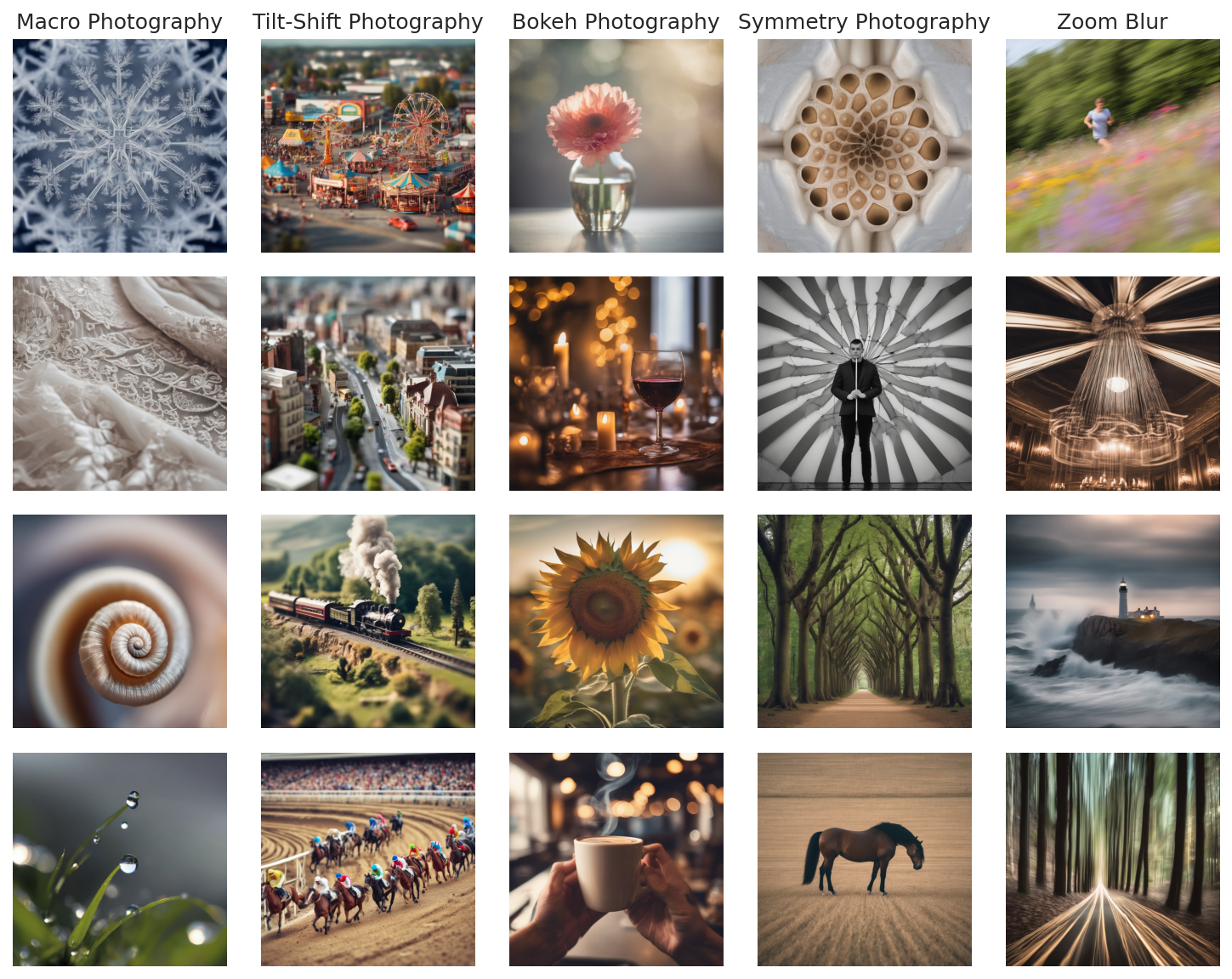}
    \vspace{-0.8em}
    \caption{Synth-Photography examples generated using Stable Diffusion XL~\cite{sdxl}, each column is a photography effect which forms the domain.}
    \label{fig:photo_grid}
\end{figure}
\vspace{-0.8em}
\begin{figure}[H]
    \centering
    \includegraphics[width=0.60\textwidth]{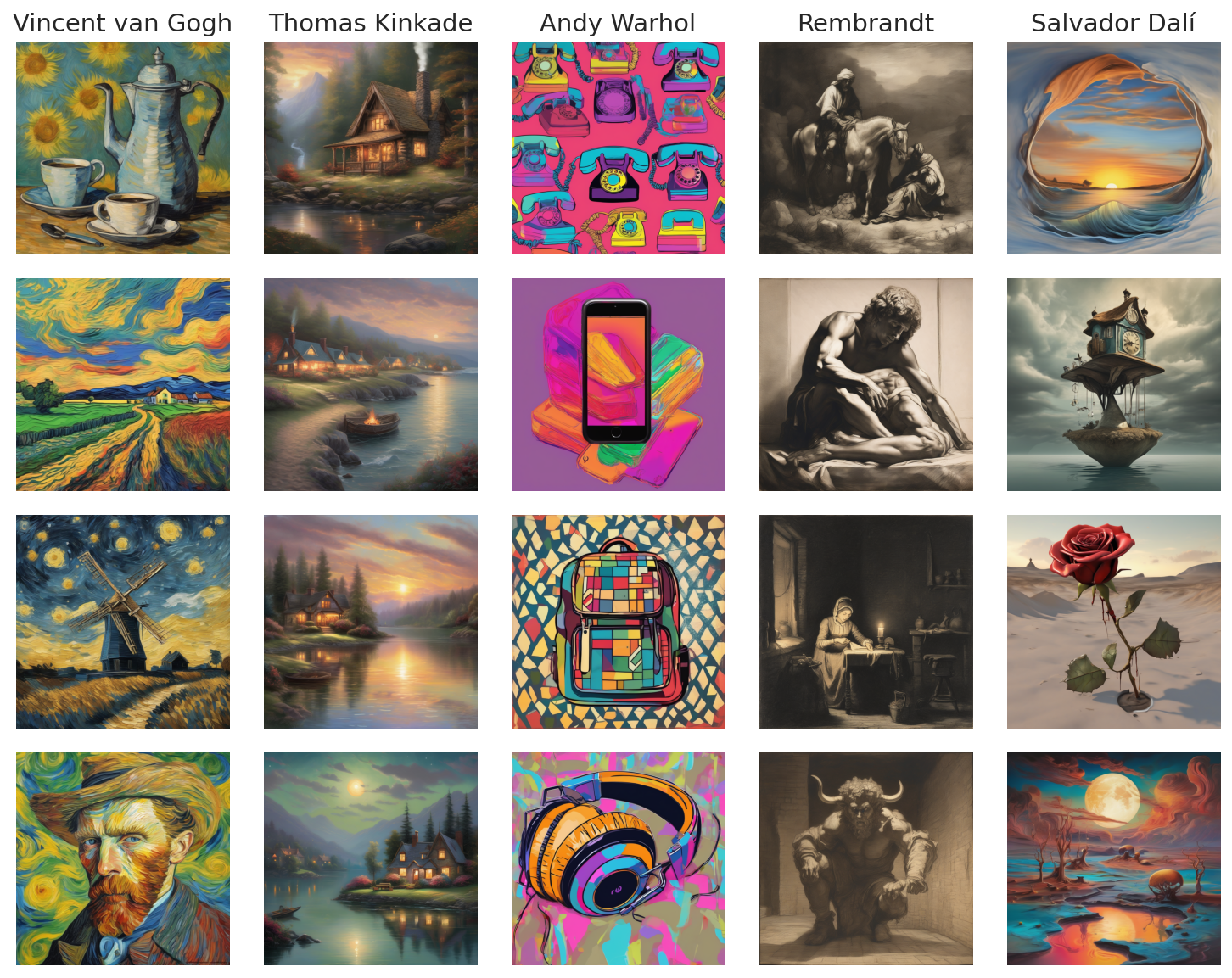}
    \vspace{-0.8em}
    \caption{Synth-Artists examples generated using Stable Diffusion XL~\cite{sdxl}, each column is an artistic style which forms the domain.}
    \label{fig:artist_grid}
\end{figure}
\vspace{-0.2em}
We generate the Synth-Photography and Synth-Artists datasets in Sec.~\ref{sec:appl} using Stable Diffusion XL~\cite{sdxl}. For Synth-photography (Fig.~\ref{fig:photo_grid}) we use the prompt \textit{``Generate an image in the style of \{effect\} photography"}; where effect can be  Macro, Tilt-Shift, Bokeh, Symmetry, and Zoom Blur. Similarly, for Synth-Artists (Fig.~\ref{fig:artist_grid}) we use the prompt \textit{``Generate an image in the style of \{artist\}"}; where artist can be Vincent Van Gogh, Thomas Kinkade, Andy Warhol, Rembrandt, and Salvador Dali.
\clearpage
\FloatBarrier


\end{document}